\pdfoutput=1
\documentclass[11pt,table]{article}
\usepackage[utf8]{inputenc}         
\usepackage[T1]{fontenc}            
\usepackage{libertine}              
\usepackage[scaled=0.85]{beramono}  
\usepackage[in]{fullpage}           
\usepackage[sc]{titlesec}           
\usepackage{xurl}                   
\usepackage{amsmath}                
\usepackage{booktabs}               
\usepackage[dvipsnames]{xcolor}     
\usepackage{graphicx}               
\graphicspath{{figures/}}           
\usepackage{subcaption}             
\usepackage{microtype}              
\usepackage[square]{natbib}         
\usepackage{siunitx}                
\usepackage{xspace}
\usepackage{listings}               
\usepackage{algorithm}              
\usepackage{algorithmic}            

\usepackage[pdfusetitle]{hyperref}  
\usepackage[nameinlink]{cleveref}   
\newcommand\myshade{90}
\definecolor{mylinkcolor}{HTML}{8F510B} 
\definecolor{mycitecolor}{HTML}{3B4791} 
\definecolor{myurlcolor}{HTML}{8F7F0B}  
\hypersetup{
    linkcolor  = mylinkcolor!\myshade!black,
    citecolor  = mycitecolor!\myshade!black,
    urlcolor   = myurlcolor!\myshade!black,
    colorlinks = true,
}

\newcommand{\tinysection}[1]{

    \vspace{0.25cm}
    \noindent
    \textbf{#1.} 
}

\newcommand{\evotorch}{\texttt{EvoTorch}\xspace}

\newcommand{\doi}[1]{\href{https://www.doi.org/#1}{doi:#1}}

\definecolor{pythonKeywordColor}{rgb}{0,0,0.5}
\definecolor{pythonStringColor}{rgb}{0.5,0,0}
\definecolor{pythonCommentColor}{rgb}{0.5,0.5,0.5}
\definecolor{pythonAreaColor}{rgb}{0.93,0.93,0.93}

\lstdefinestyle{pythonstyle}{
    backgroundcolor=\color{pythonAreaColor},   
    commentstyle=\color{pythonCommentColor},
    keywordstyle=\color{pythonKeywordColor},
    stringstyle=\color{pythonStringColor},
    basicstyle=\ttfamily\footnotesize,
    breakatwhitespace=false,         
    breaklines=true,                 
    keepspaces=true,                 
    showspaces=false,                
    showstringspaces=false,
    showtabs=false,                  
    tabsize=4
}

\begin{document}
\title{EvoTorch: Scalable Evolutionary Computation in Python}
\author{
  Nihat Engin Toklu \and Timothy Atkinson
  \and Vojt\v{e}ch Micka \and Pawe\l{} Liskowski
  \and Rupesh Kumar Srivastava\\
  $ $\\
  \{engin, timothy, vojtech, pawel, rupesh\}@nnaisense.com\\
  $ $\\
  NNAISENSE
}
\maketitle

\begin{abstract}
  Evolutionary computation is an important component within various fields such as artificial intelligence research, reinforcement learning, robotics, industrial automation and/or optimization, engineering design, etc.
  Considering the increasing computational demands and the dimensionalities of modern optimization problems, the requirement for scalable, re-usable, and practical evolutionary algorithm implementations has been growing.
  To address this requirement, we present \evotorch: an evolutionary computation library designed to work with high-dimensional optimization problems, with GPU support and with high parallelization capabilities.
  \evotorch is based on and seamlessly works with the \texttt{PyTorch} library, and therefore, allows the users to define their optimization problems using a well-known API.
\end{abstract}
\section*{About this document}

This document is a technical report describing {\bf version 0.4.0} of the \evotorch evolutionary computation library. As the library is updated, corresponding updates will be made to this technical report.
Please refer to this section, which you will find at the top of any version of this technical report, to find the corresponding version of the \evotorch library to which this report refers. 

\vspace{0.1cm}

\noindent
\evotorch is publicly available at \url{https://github.com/nnaisense/evotorch/}.

\vspace{0.1cm}

\noindent All GPU-based experiments were run on an NVIDIA DGX-Station with 4 16GB Tesla V100 GPUs, 256 GB DDR4 memory and 20-core, 40-thread Intel Xeon E5-2698 v4 processor.  

\newpage


\section{Introduction}
To use the definition of~\cite{back1996evolutionary}, evolutionary computation (EC) is ``an area of computer science that uses ideas from
biological evolution to solve computational problems''.  EC has become a popular approach for optimization and reinforcement learning due in large part to its flexibility and generality: the same evolutionary search procedure can be used to solve a wide range of tasks from static traveling salesman problems to optimizing neural network control policies.  All that's required in this black box setting is  a function computes the relative "fitness" of each candidate solution, i.e. no gradient information needed.  
A the same time, when useful domain knowledge is available, EC makes it easy to incorporate this through problem-specific operators for further efficiency.

In order to leverage the full potential of this powerful approach, parallelization is the key---just as it, resoundingly, has been in the field of deep learning where today the default approach is to train enormous neural network models on GPUs~\citep{chellapilla2006,ciresan2011,krizhevsky2012} (see survey~\citep{schmidhuber2015}) using frameworks such as TensorFlow, Theano, PyTorch,  etc.~\citep{theano2010,chainer2015,tensorflow2016,yuret2016,innes2018,frostig2018,pytorch2019}.

Because evolutionary algorithms use a population of solution points that can be evaluated independently, they are naturally suited for and can significantly benefit from the parallelized computation capabilities of hardware accelerators, allowing for much larger population sizes that increase the likelihood of reliably discovering sufficiently good solutions faster.

That being said, compared to deep learning frameworks, support for hardware accelerators in EC libraries has been relatively slow in coming, as recently argued by~\citet{evojax2022}. Instead, traditional CPU-based parallelization has remained as the
standard approach (e.g. see \texttt{DEAP} \citep{deap2012}, \texttt{pymoo} \citep{pymoo2020}, \texttt{ECJ} \citep{Luke1998ECJSoftware}, \texttt{pagmo} \citep{Biscani2020}).
It is important to note that CPU-bound EC libraries do not necessarily preclude GPU acceleration.  Indeed, one can take the population provided by the library, and manually transfer it to the GPU for accelerated computation (as done in~\citet{lu2019nsga} which
uses \texttt{pymoo}).  However, such manual operations can become tricky, especially in the case of multiple GPUs, and can usually only speed up fitness evaluations, while the EC algorithm itself remains on the CPU. To better exploit the GPUs, EC libraries
based on \texttt{JAX} \citep{frostig2018}, such as \texttt{evosax}
\citep{evosax2022}, \texttt{EvoJAX} \citep{evojax2022}, and
\texttt{EvoX} \citep{huang2023} have been recently been released which can scale up to all the GPUs visible to the computer, for both fitness evaluations \emph{and} and the EC algorithm itself.

In this report, we present \evotorch, a new EC library written in Python \citep{vanrossum2009} and 
based on \texttt{PyTorch}, unlike the aforementioned \texttt{JAX}-based EC libraries.  \evotorch brings fully parallelizable EC into the well-established and rich \texttt{PyTorch} ecosystem providing more seamless compatibility with existed code-bases.
We list the design principles of \evotorch below.

\tinysection{Easy and efficient parallelization}
\evotorch allows for multiple modes of parallelization that can be used with minimal input from the user.
Like \texttt{evosax} and \texttt{EvoJAX}, \evotorch can easily parallelize computations using all the GPUs on a local computer. In many cases, the fitness function is a CPU-bound blackbox component which is impossible to transfer to GPU (e.g. a physics simulation).
For such scenarios, \evotorch supports parallelization across multiple CPUs of the same computer, or across the CPUs of multiple computers in a cluster (with the help of the \texttt{ray} \citep{ray2018} library).
Combining all, for challenging tasks which are transferable to GPU, \evotorch supports parallelization across multiple GPUs of multiple computers of a cluster as well.

\tinysection{PyTorch ecosystem}
\texttt{PyTorch} provides convenient user-interfaces and easily-extendable tools for a variety of machine learning tasks. As a result, a broad ecosystem has developed, with many powerful \texttt{PyTorch}-based libraries now powering the state-of-the-art across many domains, for example: \texttt{PyTorch Geometric} for geometric deep learning \citep{fey2019fast}, \texttt{Detectron2} for object detection \citep{wu2019detectron2}, \texttt{AllenNLP} for natural language processing \citep{Gardner2017AllenNLP}, and \texttt{TorchDrug} for drug discovery \citep{zhu2022torchdrug}. By building upon the \texttt{PyTorch} framework, \evotorch immediately provides these growing communities with a way to incorporate GPU-accelerated evolutionary computation into their research. Additionally, this decision equips the evolutionary algorithm community with direct ways to pursue research and applications in neuroevolution within the context of the extensive support for neural networks provided through the \texttt{PyTorch} ecosystem. 

\tinysection{Generality}
While \evotorch expresses numeric solutions as \texttt{PyTorch} tensors for vectorized and hardware-accelerated operations, it also allows for the evolution of variable-length lists, tuples, dictionaries which can contain tensors or other containers.
We argue that this feature allows  \evotorch to be used on a widest possible variety of optimization problems.

\section{The proposed software: EvoTorch}

\subsection{The general structure}
\evotorch consists of the following main components.

\tinysection{Problem}
A \texttt{Problem} object is where the users define their fitness functions, objective senses (i.e. minimization or maximization), number of objectives, solution structure (where the declared structure can dictate that the solutions are vectors or variable-length lists or dictionaries, etc.), data types (e.g. \texttt{torch.float32} for 32-bit floating point numbers), as well as device of computation (e.g. "cpu" or "cuda"). Additionally, for parallelized evaluation of the solutions, one can specify the number of actors (helper subprocesses in \texttt{ray}) so that the \texttt{Problem} object will spawn that many actors and split the workload of fitness evaluation among those actors. For multi-GPU usage, one can also specify the number of GPUs to be used by each actor.

A \texttt{Problem} object also serves as a toolbox for the programmer. It provides utilities for evaluating solutions (abstracting away details regarding parallelization) and for generating populations, where each population is represented by a \texttt{SolutionBatch} that provides further utilities to the programmer for selecting solutions and/or for manipulating the population.

\tinysection{SolutionBatch}
A \texttt{SolutionBatch} is how populations and sub-populations are represented in \evotorch, storing both the decision values and the evaluation results (i.e. fitnesses and optionally further evaluation-related data) of the solutions.

If the problem at hand is configured to work with fixed-length numeric vectors, the decision values are stored in a contiguous 2-dimensional PyTorch tensor (each row of the tensor representing a different solution).
If the problem has a custom solution structure (e.g. variable-length lists, dictionaries, etc.), then the decision values are stored by an \evotorch-specific container named \texttt{ObjectArray}, which is an array type with a torch-inspired interface but with the capability of storing non-numeric data.

\texttt{SolutionBatch} provides utilities for sorting, pareto-sorting, and ranking its contained solutions, abstracting away details such as the objective sense. A \texttt{SolutionBatch} can be indexed (to get a single solution), sliced (to get a sub-population), or combined with another \texttt{SolutionBatch}. These high level functionalities operate simultaneously on the solutions and their registered fitnesses, saving the programmer from manual work and from possible mistakes.

\tinysection{SearchAlgorithm}
The implementation of an EC algorithm is expected to inherit from the super-class named \texttt{SearchAlgorithm}. \evotorch provides the following ready-to-use EC algorithms: exponential and separable variations of natural evolution strategies (XNES \citep{glasmachers2010} and SNES \citep{schaul2011}), policy gradients with parameter-based exploration (PGPE \citep{sehnke2010}), cross entropy method (CEM \citep{rubinstein1997,rubinstein1999}), covariance matrix adaptation evolution strategy (CMA-ES \citep{hansen2001}), general genetic algorithms \citep{holland1975,holland1992}, non-dominated sorting genetic algorithm (NSGA-II \citep{deb2002}), cooperative synapse neuroevolution (CoSyNE \citep{gomez2008accelerated}), and multi-dimensional archive of phenotypic elites (MAP-Elites \citep{mouret2015illuminating}). 

Users can develop custom EC algorithms via subclassing \texttt{SearchAlgorithm}. Thanks to the utilities provided by the \texttt{Problem} and \texttt{SolutionBatch} components, users do not have to worry about details such as parallelization, objective sense, etc., and can instead focus on the core algorithm implementation.

\tinysection{Logger}
\evotorch provides loggers which periodically report/save the current status of the evolutionary search to the screen, or to the disk, or to another back-end such as \texttt{neptune} \citep{neptune}, \texttt{mlflow} \citep{mlflow}, or \texttt{sacred} \citep{greff2017}.

\subsection{Library interface}
\label{sec:interface}

\evotorch is designed to work with arbitrary fitness functions which map a PyTorch tensor, representing solutions, to another PyTorch tensor, representing the evaluation results or fitnesses.

As an example, consider a function $f$ which computes the Euclidean norm of a vector $x$, and returns it as the fitness of $x$: 
\begin{lstlisting}[language=Python]
    import torch
    def f(x: torch.Tensor) -> torch.Tensor:
        return torch.linalg.norm(x)
\end{lstlisting}
Let us now assume that the optimization problem has the following properties:
(\textbf{p1}) the goal is to minimize $f(x)$;
(\textbf{p2}) a vector $x\in R^{100}$ (i.e.\ 100 decision variables);
(\textbf{p3}) the decision variables are of the type \texttt{float32};
and
(\textbf{p4}) the search should begin from within the interval $x_i\in[-10, 10], \forall i=1..100$.
In \evotorch, this optimization problem can be expressed as:
\begin{lstlisting}[language=Python]
    from evotorch import Problem
    prob = Problem(
        "min", f,             # (p1)  Note: "max" is also supported for maximization
        solution_length=100,  # (p2)
        dtype=torch.float32,  # (p3)  Note: float32 is the default, so, this line can be omitted
        initial_bounds=(-10.0, 10.0),  # (p4)
    )
\end{lstlisting}
Let us further assume, for the sake of this example, that we wish to solve the optimization problem \texttt{prob} via the eross entropy method (CEM; \citep{rubinstein1997,rubinstein1999}), and to tune it as follows:
(\textbf{c1}) the initial standard deviation of the search distribution is 0.5;
(\textbf{c2}) the population size is 250;
and
(\textbf{c3}) at each generation, the better half of the population is selected to produce the next generation's center point.
In \evotorch, CEM can be instantiated and tuned as follows:
\begin{lstlisting}[language=Python]
    from evotorch.algorithms import CEM

    searcher = CEM(
        prob,
        stdev_init=0.5,        # (c1)
        popsize=250,           # (c2)
        parenthood_ratio=0.5,  # (c3)
    )
\end{lstlisting}
This CEM instance can then be executed for 100 generations as follows:
\begin{lstlisting}[language=Python]
    from evotorch.logging import StdOutLogger

    _ = StdOutLogger(searcher)  # This will print out the status at the end of each generation
    searcher.run(100)  # Run the search algorithm for 100 generations
    result = searcher.status["center"]  # Get the center solution
\end{lstlisting}
where the solution result is expressed as the center of the evolutionary search distribution, stored in the variable \texttt{result}.

\subsection{Vectorization capabilities}

The example shown in section~\ref{sec:interface} can be made more efficient by replacing the fitness function with its \emph{vectorized} counterpart. In this context, a vectorized function is a function that operates not on a single input, but on a batch of inputs.
Recall from the example that the function $f$ expects 1-dimensional PyTorch tensors of length 100. The vectorized counterpart of $f$ instead would receive a 2-dimensional tensor of size $(k, 100)$, and would return a 1-dimensional tensor of length $k$, where $k$ is the batch size.
The vectorized counterpart of $f$ can be defined as follows:
%
%
\begin{lstlisting}[language=Python]
    from evotorch.decorators import vectorized

    @vectorized
    def vf(x: torch.Tensor) -> torch.Tensor:
        return torch.linalg.norm(x, dim=-1)
\end{lstlisting}
In this new function definition, the configuration \texttt{dim=-1} tells the PyTorch function \texttt{norm} that the norms should be computed across the last dimension. Given that the expected input is 2-dimensional, the result of this operation will be a 1-dimensional vector. This function is marked by using \texttt{@vectorized} to inform \evotorch that this function is vectorized.
After these changes, the instantiation of \texttt{Problem} is as before:
\begin{lstlisting}[language=Python]
    vprob = Problem("min", vf, ...)
\end{lstlisting}
%
%
%
Vectorized PyTorch functions are suitable for further performance boost from GPUs.
In \evotorch, it is possible to move the entire evolutionary procedure (including the search algorithm and the fitness function) to the GPU by simply using the \texttt{device} keyword argument:
%
%
%
\begin{lstlisting}[language=Python]
    vprob = Problem("min", vf, device="cuda:0", ...)
\end{lstlisting}
where \texttt{"cuda:0"} refers to the first GPU device usable by the CUDA backend of PyTorch.

\subsection{Parallelization capabilities}

In some scenarios, one might prefer traditional CPU-based multiprocess parallelization over GPU-based vectorization. One common scenario is when the fitness function is not (or cannot) be defined purely via PyTorch operations.
For such cases, \evotorch has the capability of instantiating \emph{actors} (with the help of the \texttt{ray} framework~\citep{ray2018}), each actor being a separate process dedicated to evaluating the fitness of the solutions it receives. The performance boost is then realized by sharing a population among the actors and running them in parallel.
Enabling such parallelization is done using the keyword argument \texttt{num\_actors}:
\begin{lstlisting}[language=Python]
    parallel_prob = Problem("min", f, num_actors=4, ...)  # use 4 actors in parallel
\end{lstlisting}
%

\subsection{Combining parallelization and GPU vectorization}

CPU-based parallelization and GPU vectorization can be combined to take advantages of multiple GPUs. In those cases, one can define the main device of the problem as "cpu", and use a special decorator named \texttt{@on\_cuda} to mark the fitness function so that each actor will use the CUDA device assigned to itself when calling the fitness function.
An example usage of the decorator \texttt{@on\_cuda} is shown below:
\begin{lstlisting}[language=Python]
    from evotorch.decorators import on_cuda, vectorized

    @on_cuda
    @vectorized
    def gpu_f(x: torch.Tensor) -> torch.Tensor:
        fitnesses = ...  # compute each fitness somehow
        return fitnesses
\end{lstlisting}
Now, assuming four GPUs, the following problem instantiation can be made:
\begin{lstlisting}[language=Python]
    multi_gpu_prob = Problem("min", gpu_f, num_actors=4, num_gpus_per_actor=1, device="cpu", ...)
\end{lstlisting}
where the main device for the \texttt{Problem} object is "cpu" (meaning that populations will be held on the CPU, and the evolutionary algorithm will run its operations on the CPU), but once a remote actor receives its portion of the population for fitness evaluation, it will move that portion onto its assigned GPU and execute the fitness function there.
The argument \texttt{num\_actors=4} specifies that there will be four actors, and \texttt{num\_gpus\_per\_actor=1} specifies that each actor will be assigned its own GPU.

Optionally, the algorithms CEM, PGPE, and SNES can further exploit multiple GPUs by enabling a \emph{distributed} mode inspired by the distributed evolutionary search architectures used in \citep{salimans2017,mania2018,lenc2019non}.
In the distributed mode, the following steps are followed at each generation:
(\emph{i}) the main process sends each remote actor its current search distribution parameters;
(\emph{ii}) each remote actor generates its own sub-population according to these parameters on its own assigned fitness evaluation device (which, in the case of the example function \texttt{gpu\_f}, is its assigned "cuda" device);
(\emph{iii}) each remote actor evaluates its own sub-population, and computes its own sub-gradient;
(\emph{iv}) the main process collects the sub-gradients from the remote actors and 
computes the main gradient via averaging;
and (\emph{v}) the main gradient is used for updating the parameters of the search distribution.
Thanks to the step \emph{(ii)}, the generation of the population is handled in a parallelized manner (in addition to the parallelization applied on fitness evaluation).
Furthermore, the communication between the main process and the remote actors is reduced to the search distribution parameters and the sub-gradients.
Therefore, further performance gains can be observed in distributed mode. 
The distributed mode can be enabled simply via the keyword argument \texttt{distributed}, as shown below:
\begin{lstlisting}[language=Python]
    searcher = PGPE(multi_gpu_prob, distributed=True, ...)
\end{lstlisting}

\section{Use Cases}

To highlight the general applicability of \evotorch and the various research directions it enables, a number of example use-cases are provided below. 

\subsection{Blackbox Optimization}

\begin{figure}[!t]
    \centering
    \includegraphics[width=0.6\textwidth]{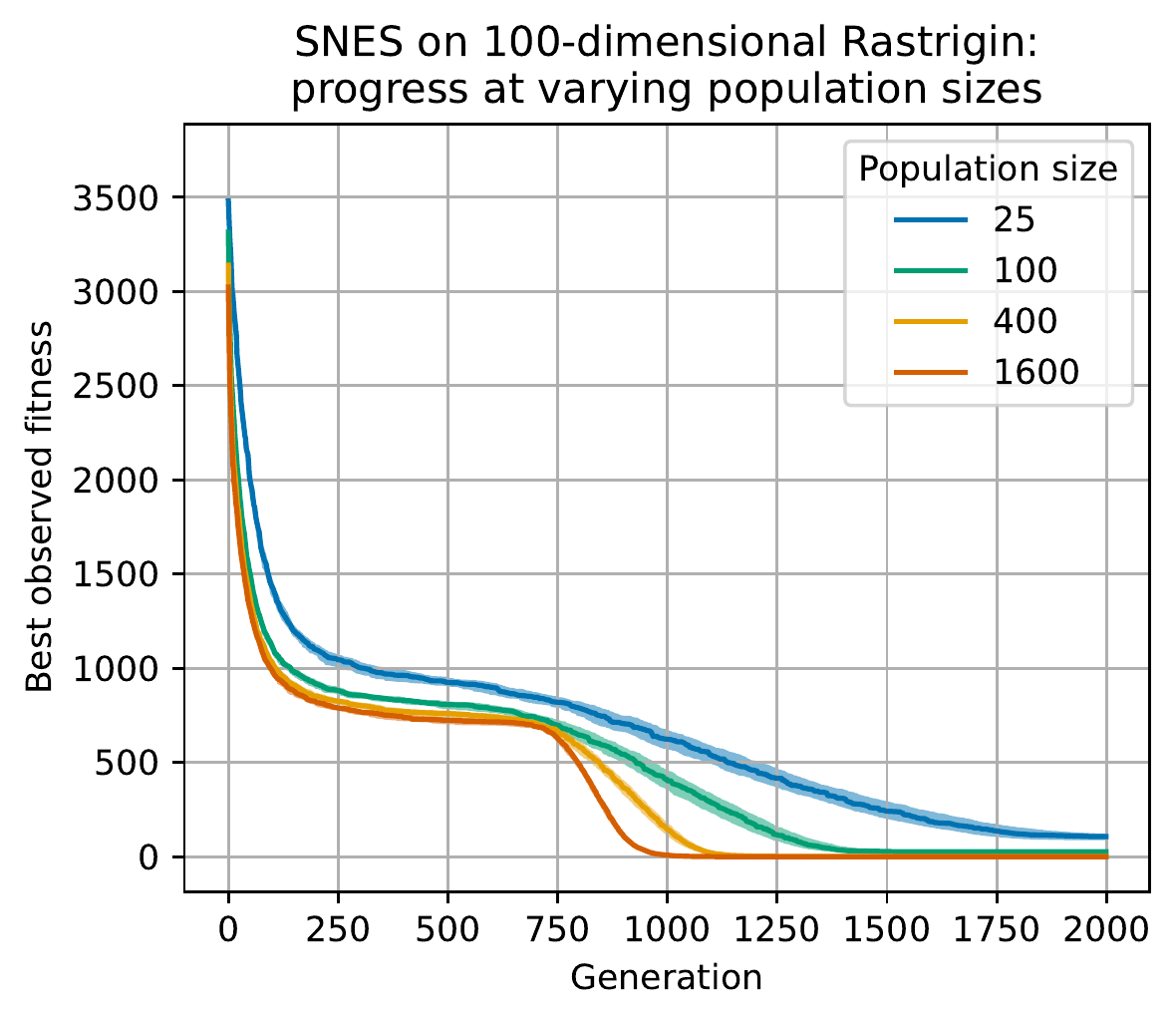}
    \caption{{\bf Comparison of the median fitness of the best solution for different population size using SNES on the 100-dimensional Rastrigin function.} Bold curves represent the medians over 50 runs, and the shaded regions represent the means $\pm$ the standard deviations.}
    \label{fig:snes_rastrigin_popsize}
\end{figure}

An immediate application of \evotorch is general blackbox optimization problems. Consider the classic Rastrigin blackbox optimization benchmark function~\citep{hoffmeister1990genetic} over $n$ dimensions,

\begin{equation}
    f(x) = 10 n + \sum_{i=1}^n x_i^2 - 10 \cos(2 \pi x_i),
\end{equation}

which can be implemented in vectorized form as,

\begin{lstlisting}[language=Python]
    @vectorized
    def vrast(x: torch.Tensor) -> torch.Tensor:
        return 10 * x.shape[-1] + torch.sum(x.pow(2.) - 10 * torch.cos(2 * math.pi * x), dim=-1)
\end{lstlisting}

which gives the vectorized and GPU-ready minimization problem,

\begin{lstlisting}[language=Python]
    rastrigin_prob = Problem(
        "min", vrast, device="cuda:0", solution_length=100, initial_bounds=(-5.12, 5.12))
\end{lstlisting}

It is then straightforward to run a continuous blackbox optimization algorithm such as SNES~\citep{schaul2011} to optimize this task,

\begin{lstlisting}[language=Python]
    searcher = SNES(rastrigin_prob, stdev_init=5)
    searcher.run(4000)  # Run for 4000 generations
\end{lstlisting}.

However, note that the main measure of algorithm performance in blackbox benchmarking literature is the fitness obtained after a given budget of samples~\citep{hansen2010comparing}. The underlying assumption is that every sample has equal cost, but this is not strictly the case when optimization problems are vectorizable and parallelizable, which most, if not all, problems are. When time-to-solution or solution quality, rather than raw work done, is the most important criteria, vectorization and parallelization may fundamentally shift how algorithms are evaluated.

\begin{figure}[!t]
    \centering
    \includegraphics[width=0.6\textwidth]{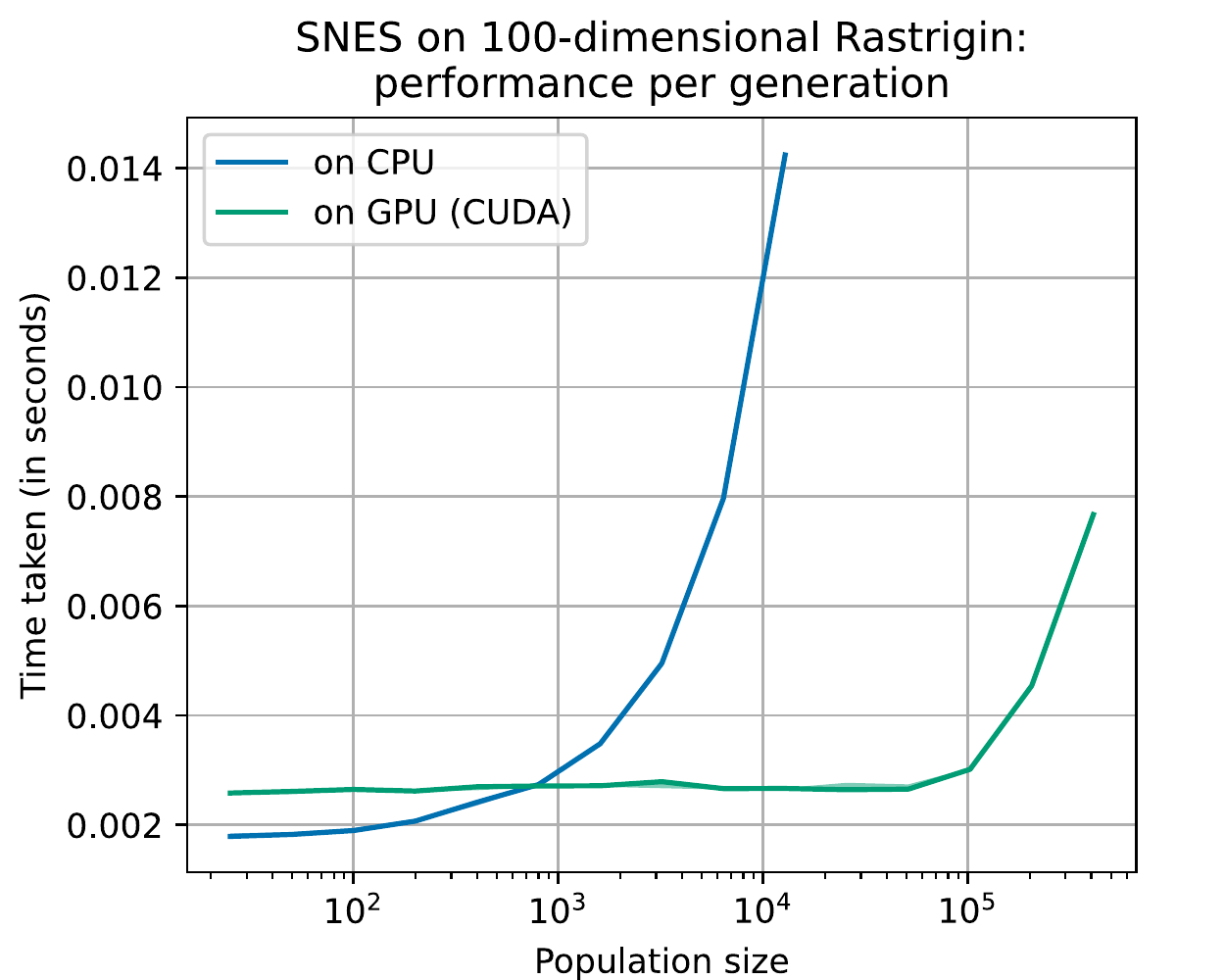}
    \caption{{\bf GPU vs CPU wall-clock time for a simple black-box optimization experiment per generation.} 
    Bold curves represent the medians, and the shaded regions represent the means $\pm$ the standard deviations.
    For each population size, 50 runs were started, each run lasting for 2000 generations. Median time was summarized over all generations of all runs. Therefore, each reported time for each population size is a summary over 2000 $\times$ 50 = 100000 data points.
    }
    \label{fig:snes_rastrigin_runtime}
\end{figure}

Figure~\ref{fig:snes_rastrigin_popsize} shows the median fitness of the best solution to the 100-dimensional Rastrigin problem discovered by SNES across 50 runs, with varying population size $\lambda \in \{25, 100, 400, 1600\}$. Substantial increases in population size visibly improve both the speed of convergence and quality of optima discovered. However, the most surprising aspect of this is the tradeoff in computation time; in Figure~\ref{fig:snes_rastrigin_runtime} we present run-time performance results for the SNES implementation on both CPU and GPU. Here we observe that as the population size increases, the implementations scale elegantly into the 10s and even (in the case of GPU) hundreds of thousands. 

\subsection{As an alternative to first-order optimization methods}

In some scenarios, one might consider using EC as an alternative to a gradient-aware first-order optimization method.
One such scenario is when the fitness function is known to be differentiable, yet its implementation is only available on a framework which does not support auto-differentiation.
Another such scenario is when, even though the fitness function is differentiable, the search space is nonconvex, and therefore there is a significant chance of converging to a suboptimal solution when relying on local gradients.

To be a practical alternative to first-order optimization methods, an evolutionary algorithm has to be competitive in terms of its finally converged solutions, and also in terms of its wall-clock time requirements.

Here, we evaluate the evolutionary algorithm CEM to see if it is indeed competitive against the well-known first-order optimization method Adam~\citep{kingma2015}. We also add a parallelized variant of Adam into our comparison, where $n$ instances of Adam perform their searches in parallel, $n$ being the population size, equal to the population size we use for CEM.

\begin{figure}[t!]
    \textbf{(a)} Sphere function

    \includegraphics[width=0.32\textwidth]{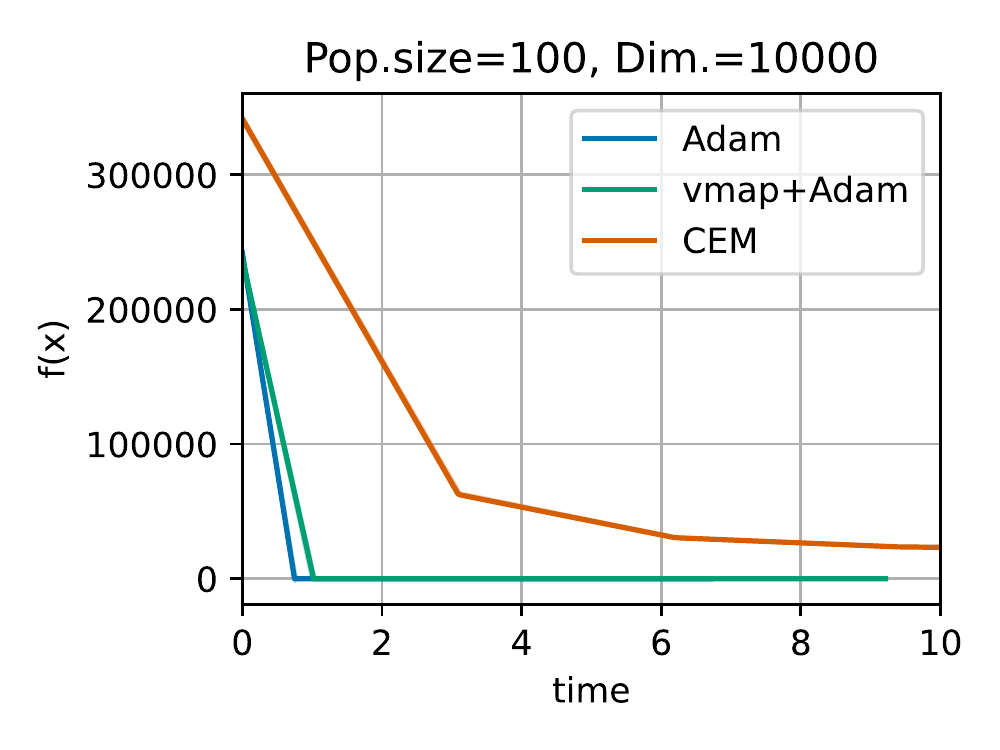}
    \includegraphics[width=0.32\textwidth]{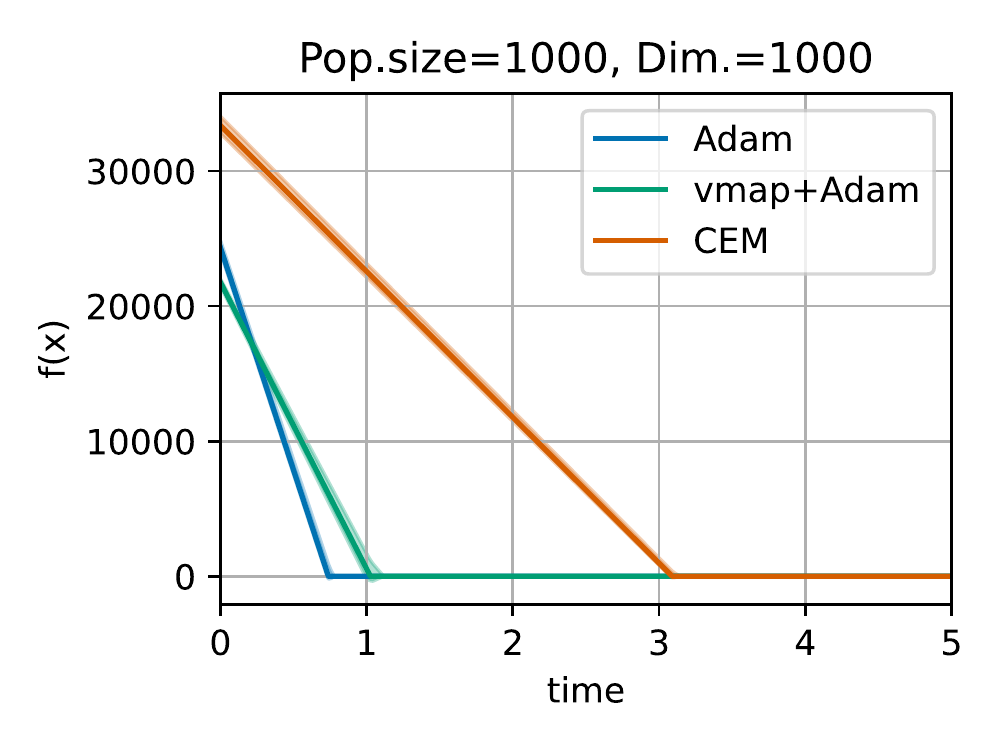}
    \includegraphics[width=0.32\textwidth]{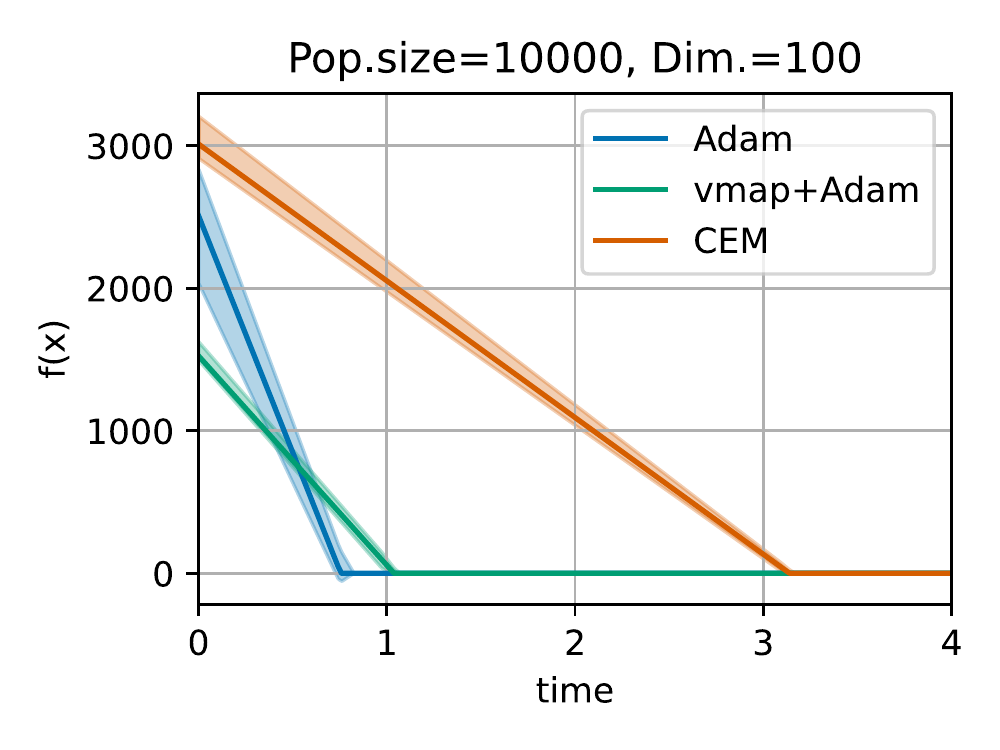}

    \textbf{(b)} Rastrigin function

    \includegraphics[width=0.32\textwidth]{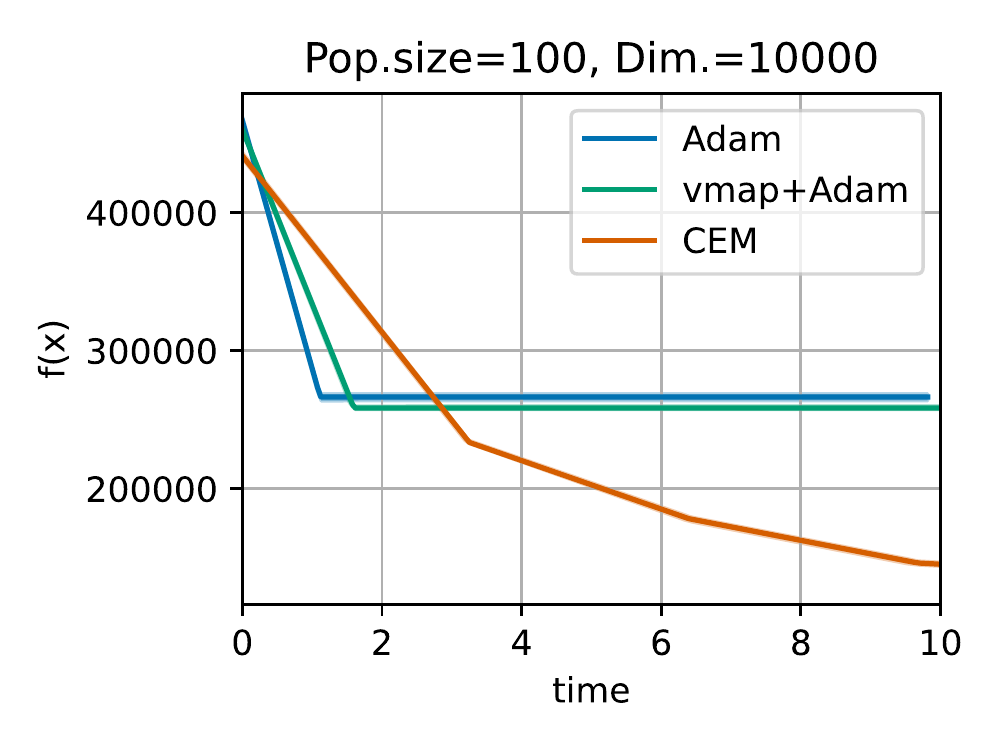}
    \includegraphics[width=0.32\textwidth]{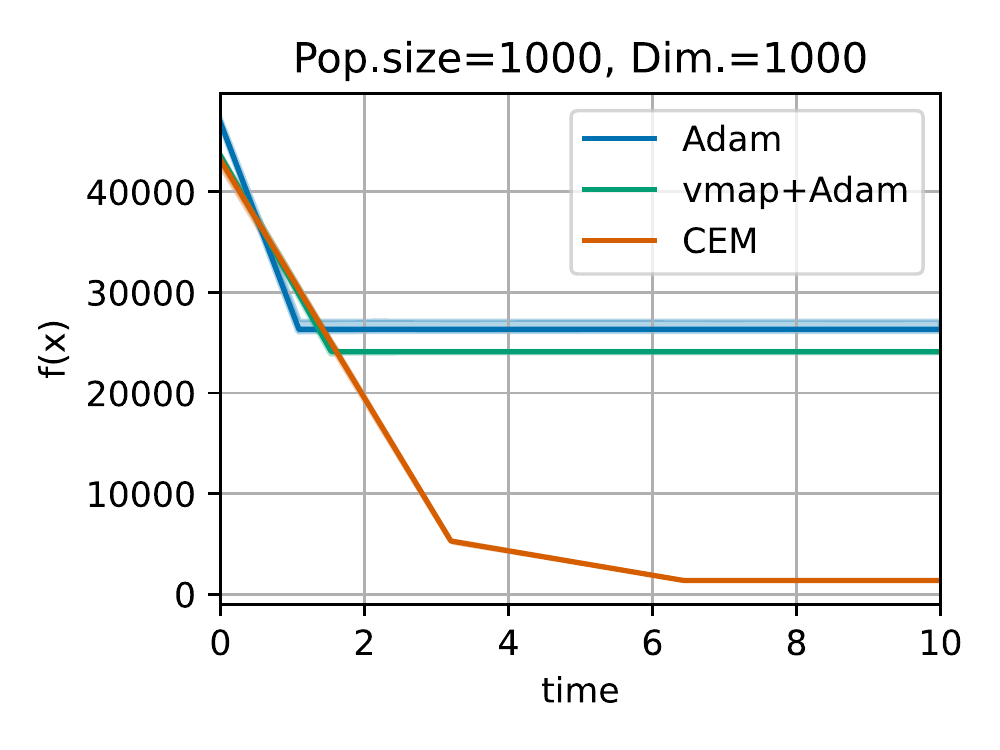}
    \includegraphics[width=0.32\textwidth]{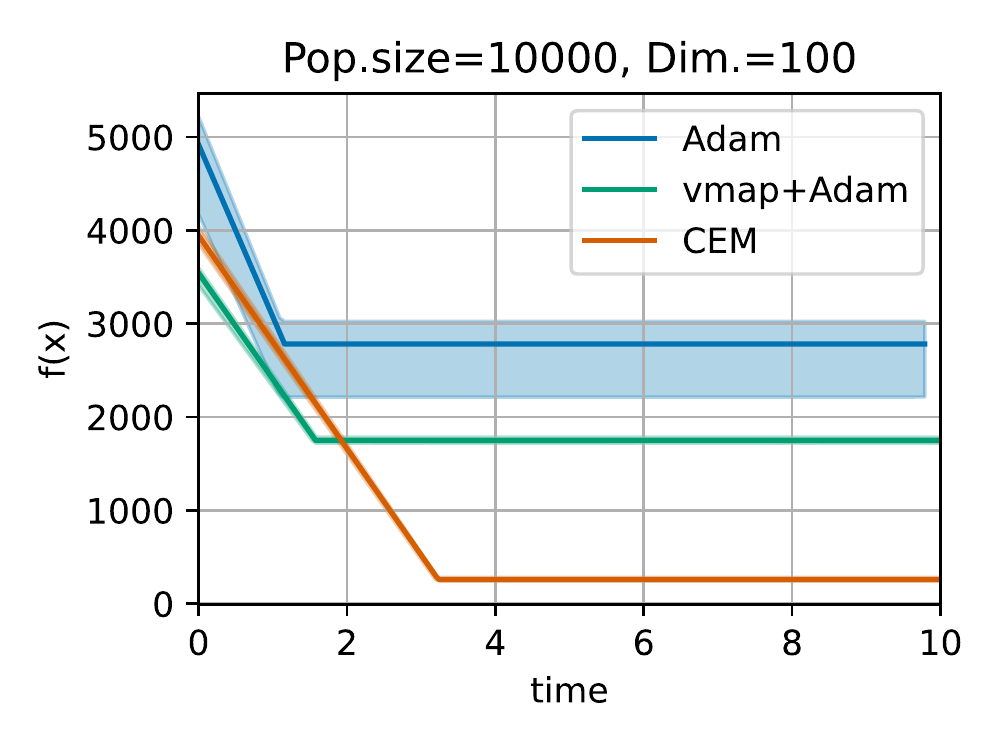}

    \begin{tabular}{p{0.9\textwidth}}
    { \footnotesize
    For both sphere and Rastrigin functions, initial search points were picked elementwise from within $[-10, 10]$.
    }
    \\
    { \footnotesize
    \underline{Hyperparameters for CEM.}
    Initial standard deviation of the search distribution: 1.0.
    Parenthood ratio: 0.5 (i.e. the better half of each population is declared as parents).
    An element of the standard deviation vector is not allowed to change more than 2\% of its original value (this limiter was previously used in the PGPE implementation of \cite{ha2019}, where 20\% was used).
    }
    \\
    { \footnotesize
    \underline{Hyperparameters for Adam.}
    Step size: 0.6 (tuned from the set $\{0.0025, 0.005, 0.01, 0.02, 0.1, 0.4, 0.6, 1.0\}$).
    Other settings are left as default ($\beta_1=0.9$, $\beta_2=0.999$, $\epsilon=\textnormal{1e-8}$).
    }
    \end{tabular}

    \caption{
    { \bf
        Convergence plots (fitness $f(x)$ vs wall clock time in seconds) comparing Adam, vmap+Adam (parallelized Adam), and CEM, on (a) sphere function and (b) Rastrigin function, considering three pairs of population sizes and problem dimensionalities.
    }
    Each curve represents the median across 5 runs.
    Shaded regions are bounded by the means $\pm$ the standard deviations.
    }
    \label{fig:vsadam}
\end{figure}

Our test problems are the convex sphere problem ($\sum_{i \in \{1,2,...\}} x^2$) and the nonconvex Rastrigin problem.
For each test problem, we used three settings:
(\emph{i}) population size=10000, dimensionality of the problem=100;
(\emph{ii}) population size=1000, dimensionality of the problem=1000; and
(\emph{iii}) population size=100, dimensionality of the problem=10000.
These three settings ensure that the computations required by each iteration for each of the considered methods can be fully parallelized across the cores of a single GPU.
This test, therefore, aims to demonstrate how EC compares against Adam when GPU-based parallelization is fully exploited.
The results are shown in figure \ref{fig:vsadam}.

In figure \ref{fig:vsadam}(a), unsurprisingly, both variations of Adam quickly converge to the optimum point of the sphere function thanks to the accurate directions provided by the true gradients. Interestingly, we also see that the time required by CEM to converge to the optimum is in the same order of magnitude compared to the single Adam run (except when its population size is 100 where it cannot reach the optimum), despite the fact that CEM has to evaluate many test points at each iteration. This result demonstrates a case where GPU-based parallelization brings the performance of EC to a level that is comparable to first-order methods even on simple convex functions.

In figure \ref{fig:vsadam}(b), we see that both variations of Adam suffer from the locality of their gradients, and converge to worse local optima on the Rastrigin function compared to CEM. These results are compatible with the study of \cite{lehman2018more}, where the authors show that a first-order gradient descent struggles on the "fleeting peaks" surface, a search space which has nonconvex features.

In summary, while the accuracy of the true gradients and the usefulness of the first-order methods cannot be denied, these empirical results emphasize the generality of evolutionary algorithms, and show us that GPUs further increase this generality.

\subsection{multi-objective Optimization}

\begin{figure}[t!]
    \begin{subfigure}{.49\textwidth}
    \caption{Population size $25$}
    \centering
    \includegraphics[width=\textwidth]{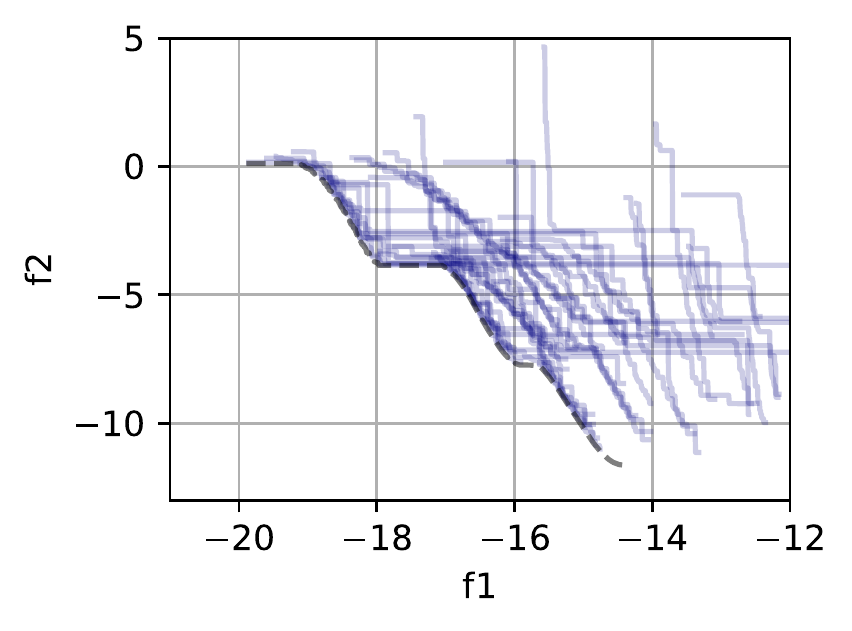}
    \end{subfigure}
    \begin{subfigure}{.49\textwidth}
    \caption{Population size $100$}
    \centering
    \includegraphics[width=\textwidth]{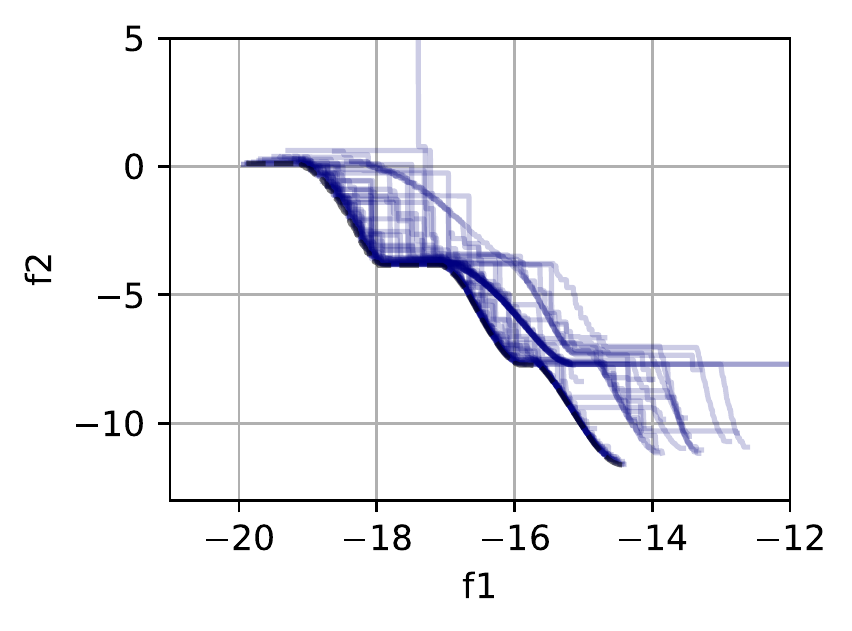}
    \end{subfigure}
    \begin{subfigure}{.49\textwidth}
    \caption{Population size $400$}
    \centering
    \includegraphics[width=\textwidth]{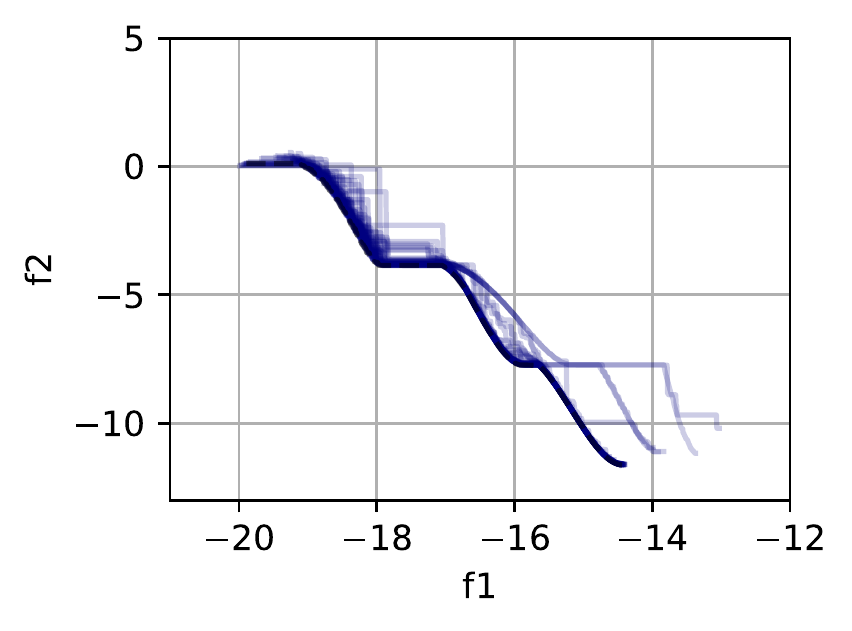}
    \end{subfigure}
    \begin{subfigure}{.49\textwidth}
    \caption{Population size $1600$}
    \centering
    \includegraphics[width=\textwidth]{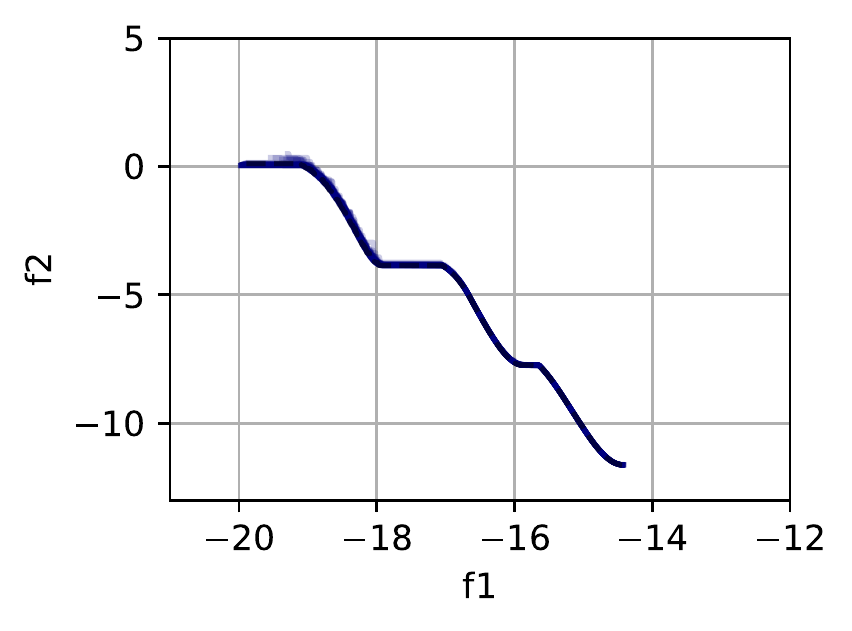}
    \end{subfigure}
    \caption{{\bf Results for NSGA-II on the Kursawe function.} Varying population sizes: (a) 25, (b) 100, (c) 400 and (d) 1600. Each faint blue line represents a Pareto front discovered by a single run of the algorithm, with 50 runs performed for each population size. The dashed black line represents the known global Pareto front of solutions.}
    \label{fig:nsga_ii_popsize}
\end{figure}

Evolutionary algorithms are, alongside mathematical programming methods, one of the most popular approaches to a-posteriori multi-objective optimization. A specific advantage of evolutionary algorithms in this setting is that they can use membership of the Pareto front, and heuristics thereof, as a selection criteria to evolve a diverse population of candidate solutions without prior knowledge of how to weight the multiple, potentially conflicting, rewards. As a result, multi-objective evolutionary algorithms such as the NSGA family~\citep{deb2002,deb2013evolutionary}, MOEA/D~\cite{zhang2007moea} and AGE-MOEA~\citep{panichella2019adaptive} have found widespread application, for example in planning~\citep{sarker2009improved}, molecular docking~\citep{janson2008molecular} and neural architecture search~\citep{lu2019nsga}.

\evotorch facilitates the straight-forward implementation and application of multi-objective optimization at scale. By default, \evotorch problem definitions support multiple objectives. Consider the bi-objective, tri-variate Kursawe function \citep{kursawe1991variant},

\begin{equation}
    f_1(x) = \sum_{i=1}^2 -10 e^{-0.2 \sqrt{x_i^2 + x_{i+1}^2}},
\end{equation}

\begin{equation}
    f_2(x) = \sum_{i=1}^3 |x_i|^{0.8} + 5 \sin x_i^3.
\end{equation}

This function can be easily implemented in vectorized form using PyTorch such that the declared function receives a $(k, 3)$ dimensional tensor of $k$ 3-dimensional vector solutions, and returns a $(k, 2)$ dimensional tensor of $k$ 2-dimensional fitnesses:

\begin{lstlisting}[language=Python]
    @vectorized
    def kursawe(x: torch.Tensor) -> torch.Tensor:
        f1 = torch.sum(
            -10 * torch.exp(
                -0.2 * torch.sqrt(x[:, 0:2] ** 2.0 + x[:, 1:3] ** 2.0)
            ),
            dim=-1,
        )  # Compute the first objective
        f2 = torch.sum(
            (torch.abs(x) ** 0.8) + (5 * torch.sin(x ** 3)),
            dim=-1,
        )  # Compute the second objective
        fitnesses = torch.stack([f1, f2], dim=-1)  # Stack the two objectives together
        return fitnesses
\end{lstlisting}
Then a problem class can be instantiated as normal,
\begin{lstlisting}[language=Python]
    kursawe_prob = Problem(
        ["min", "min"], kursawe,             # Both objectives are to be minimized
        solution_length=3, 
        initial_bounds=(-5.0, 5.0),
        solution_length=3,
    )
\end{lstlisting}
The genetic algorithm implementation of \evotorch can handle multiple objectives, and therefore can work on \texttt{kursawe\_prob}:
\begin{lstlisting}[language=Python]
    ga = GeneticAlgorithm(kursawe_prob, popsize=...)
\end{lstlisting}
When initialized with a multi-objective problem, \texttt{GeneticAlgorithm} follows the procedures of NSGA-II \citep{deb2002} and enables the behavior of organizing its populations using Pareto-ranking and crowd-sorting.

Figure \ref{fig:nsga_ii_popsize} shows the results of running this multi-objective algorithm for $30$ generations with varying population sizes. Each faint line represents the Pareto front of discovered solutions by the algorithm, with 50 runs performed for each population size. Even though the total number of generations remains the same, it can clearly be seen that increasing the population size yields substantially better Pareto fronts, with the $1600$ population size reliably obtaining the known global optimum for this problem, represented as a black dashed line in each plot. 

While scaling the population is clearly beneficial for NSGA-II, in principle, many multi-objective evolutionary algorithms require $O(n^2)$ time complexity where $n$ is the population size, as they require pairwise comparisons of the individual solutions' fitness values to establish Pareto dominance. In practice, this computationally limits the application of MOEAs to relatively small population sizes. However, the underlying mathematics of computing population-wide pareto dominance are highly vectorizable and therefore well-suited to execution on accelerated hardware such as GPUs. 

\begin{figure}[!t]
    \centering
    \includegraphics[width=0.6\textwidth]{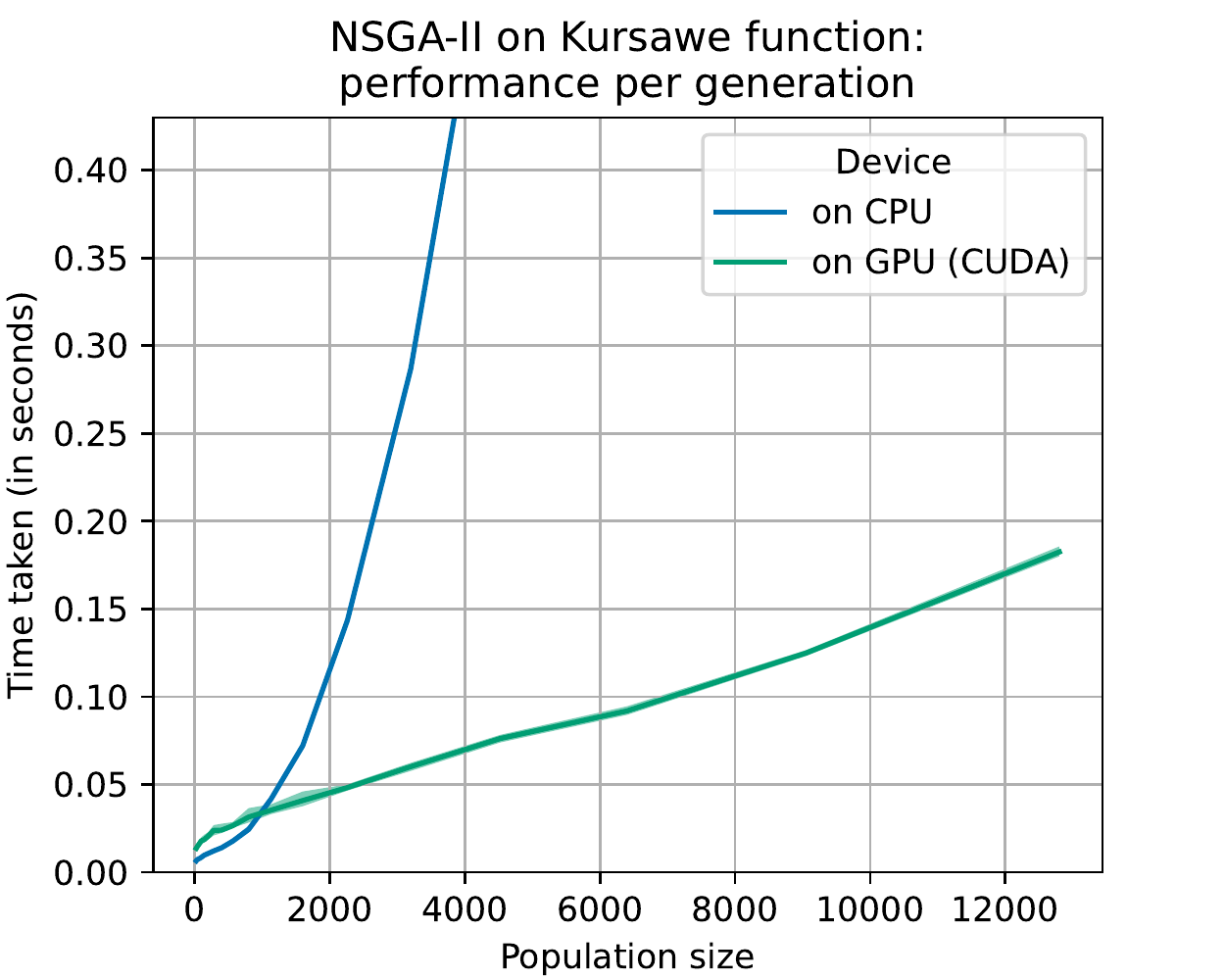}
    \caption{{\bf GPU vs CPU wall-clock times for a simple multi-objective optimization experiment per generation.}
    Bold curves represent the medians, and the shaded regions represent the means $\pm$ the standard deviations.
    For each population size, 50 runs were started, each run lasting for 30 generations. Median time was summarized over all generations of all runs. Therefore, each reported time for each population size is a summary over 30 $\times$ 50 = 1500 data points.
    }
    \label{fig:nsga_ii_runtime}
\end{figure}

As a result, when excluding the time complexity of fitness evaluation which may vary with application, we generally observe that the run-time of EvoTorch's NSGA-II implementation scales elegantly with the population size when deployed on such accelerated hardware. This is demonstrated in Figure \ref{fig:nsga_ii_runtime}, where NSGA-II is run on the Kursawe problem at varying population sizes on both the CPU and GPU. In this setting, we observe that we are able to efficiently run NSGA-II on the GPU with population sizes greater than $10000$, thanks to the fast implementation of Pareto sorting.

\subsection{Reinforcement learning}

It has been demonstrated that evolutionary algorithms are competitive for solving reinforcement learning tasks
(see, e.g., \citep{salimans2017,mania2018,ha2019}).
\evotorch provides problem types for tackling \texttt{gym} and \texttt{brax} tasks.
Additionally, utilities which were demonstrated by \cite{salimans2017} as helpful for evolutionary reinforcement learning, such as online observation normalization and adaptive population size, are also implemented.

\subsubsection{Gym}

The \texttt{gym} library \citep{brockman2016openai} provides various reinforcement learning tasks that have been used as common benchmarks. Among the environments provided by \texttt{gym} are continuous locomotion tasks (such as \emph{hopper} \citep{erez2011}, \emph{walker}, \emph{humanoid} \citep{tassa2012}, etc.) based on the \texttt{MuJoCo} simulator \citep{todorov2012mujoco}. Additional \texttt{gym} environments can be provided by libraries such as \texttt{PyBullet} \citep{coumans2021}, \texttt{Roboschool} \citep{klimov2017}, \texttt{PyBullet Gymperium} \citep{benelot2018}, etc.

\evotorch has a problem type named \texttt{GymNE} for solving reinforcement learning tasks defined by the \texttt{gym} library. A simple instantiation of \texttt{GymNE} for solving the cart pole task looks like this:
\begin{lstlisting}[language=Python]
    from evotorch.neuroevolution import GymNE
    problem = GymNE(env="CartPole-v1", network=network_class)
\end{lstlisting}
where \texttt{network\_class} is a reference to a class that inherits from \texttt{torch.nn.Module}, which specifies the architecture of the neural network whose parameters will be evolved. More keyword arguments are available for enabling observation normalization, multi-episode evaluation, hard-coded alive bonus removal (which is reported to help with standard locomotion benchmarks \citep{mania2018,toklu2020clipup}), timestep-dependent manual alive bonus schedule, parallelized evaluation (using multiple actors via \texttt{num\_actors}), etc.

\begin{figure}[!t]
    \centering
    \includegraphics[width=0.32\textwidth]{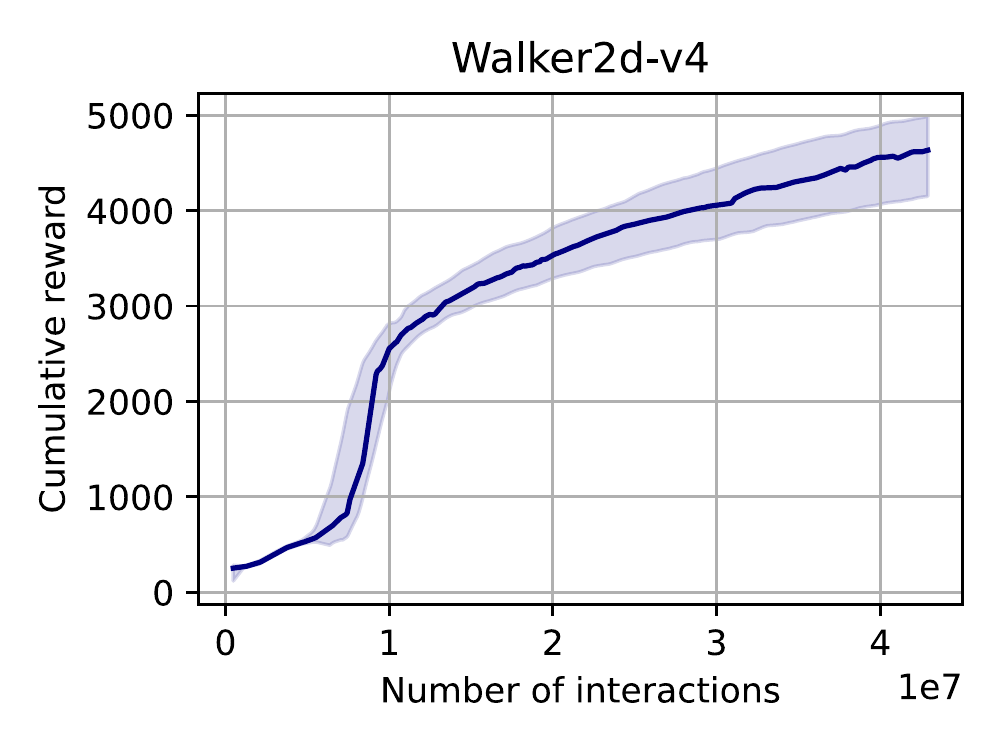}
    \includegraphics[width=0.32\textwidth]{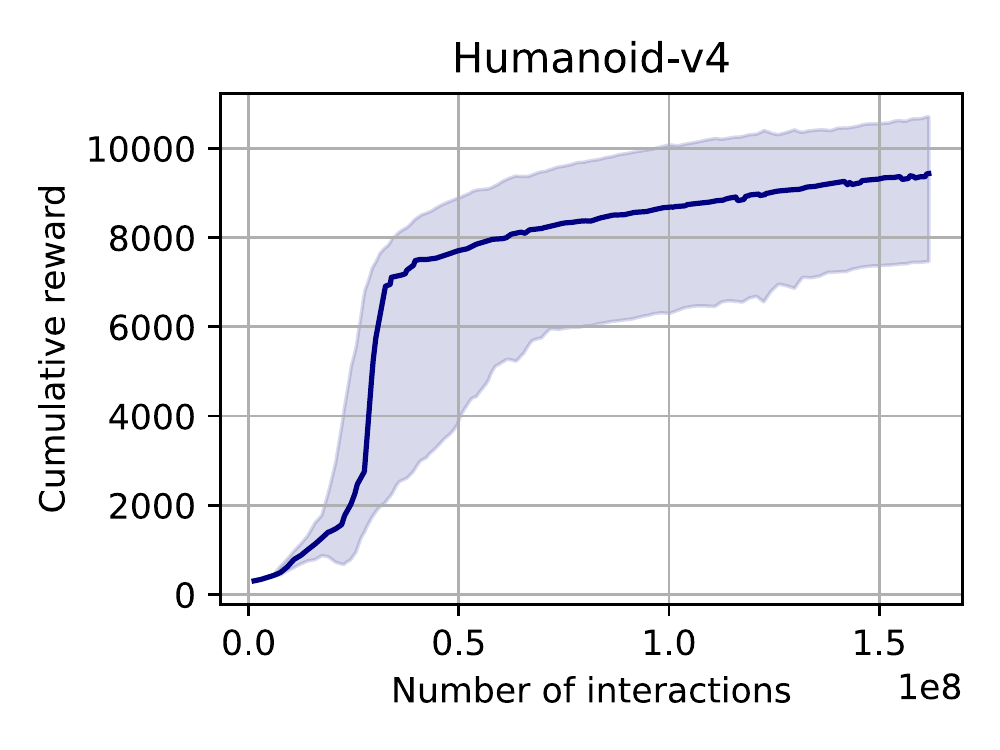}
    \includegraphics[width=0.32\textwidth]{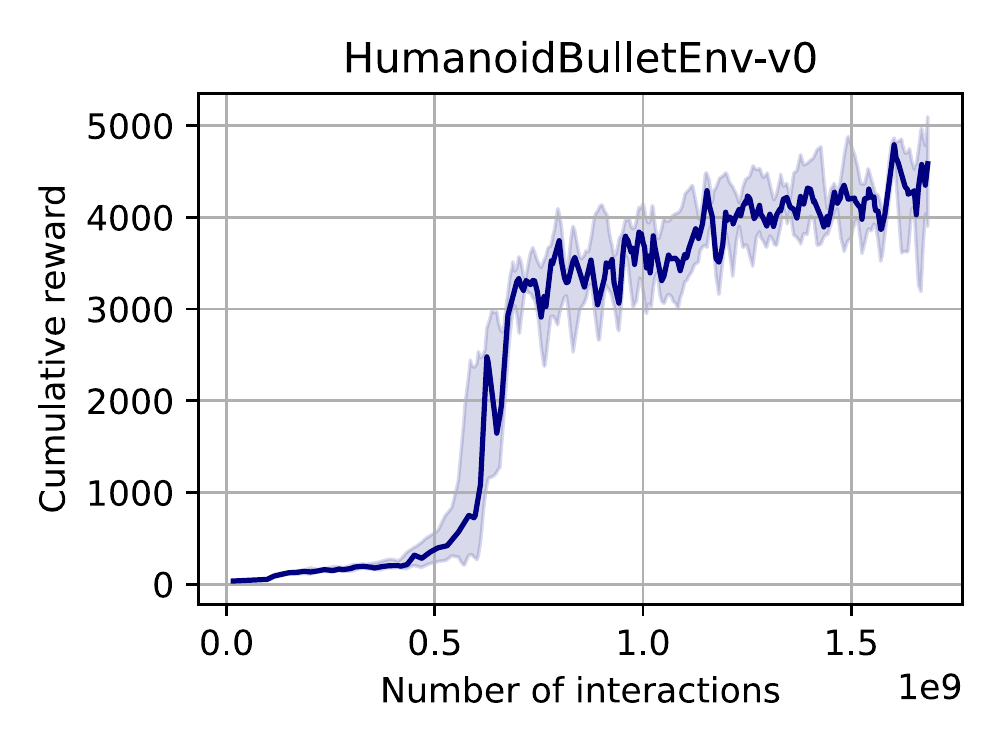}
    \\
    \begin{tabular}{p{0.9\textwidth}}
        { \footnotesize
            \underline{Configuration.}
            For \emph{Walker2d-v4} and \emph{Humanoid-v4}, linear policies with bias terms were used.
            For \emph{HumanoidBulletEnv-v0}, a neural network with a hidden recurrent layer of 64 neurons and with tanh activations was used.
            For each task, PGPE+ClipUp hyperparameters were taken from \citep{toklu2020clipup}.
            In the case of \emph{HumanoidBulletEnv-v0}, even though a non-recurrent network was used in \citep{toklu2020clipup}, same hyperparameters were used and observed to work for the runs reported here with recurrent policies.
        }
    \end{tabular}

    \caption{{\bf Results obtained on locomotion tasks with PGPE+ClipUp.}
    The plots for \emph{Walker2d-v4} and \emph{Humanoid-v4} summarize 30 runs.
    The plot for \emph{HumanoidBulletEnv-v0} summarizes 5 runs.
    In each plot, bold curve represents the median and the shaded area is bounded by the mean $\pm$ the standard deviation.
    }
    \label{fig:rlresults}
\end{figure}

It has been observed that the PGPE+ClipUp (i.e. PGPE \citep{sehnke2010} coupled with the ClipUp optimizer \citep{toklu2020clipup}) implementation of \evotorch can solve the \texttt{MuJoCo} tasks \emph{Walker2d-v4} and \emph{Humanoid-v4} with a performance that is compatible with \citep{toklu2020clipup}.
We have also observed that the \texttt{PyBullet} task \emph{HumanoidBulletEnv-v0} can be solved to obtain a competitive cumulative reward.
The results are shown in figure \ref{fig:rlresults}.


\subsubsection{Brax}

\texttt{brax} \citep{freeman2021brax,brax2021github} is a differentiable and vectorized simulator written in \texttt{JAX} that can take advantage of hardware accelerators including GPUs.
\evotorch provides a vectorized counterpart of \texttt{GymNE} named \texttt{VecGymNE}, that can work with \texttt{brax} tasks.

\begin{figure}[!p]
    \centering

    \begin{tabular}{p{0.9\textwidth}}
        \textbf{(a)} fetch
        \\
        \begin{tabular}{c c}
            \includegraphics[width=0.4\textwidth]{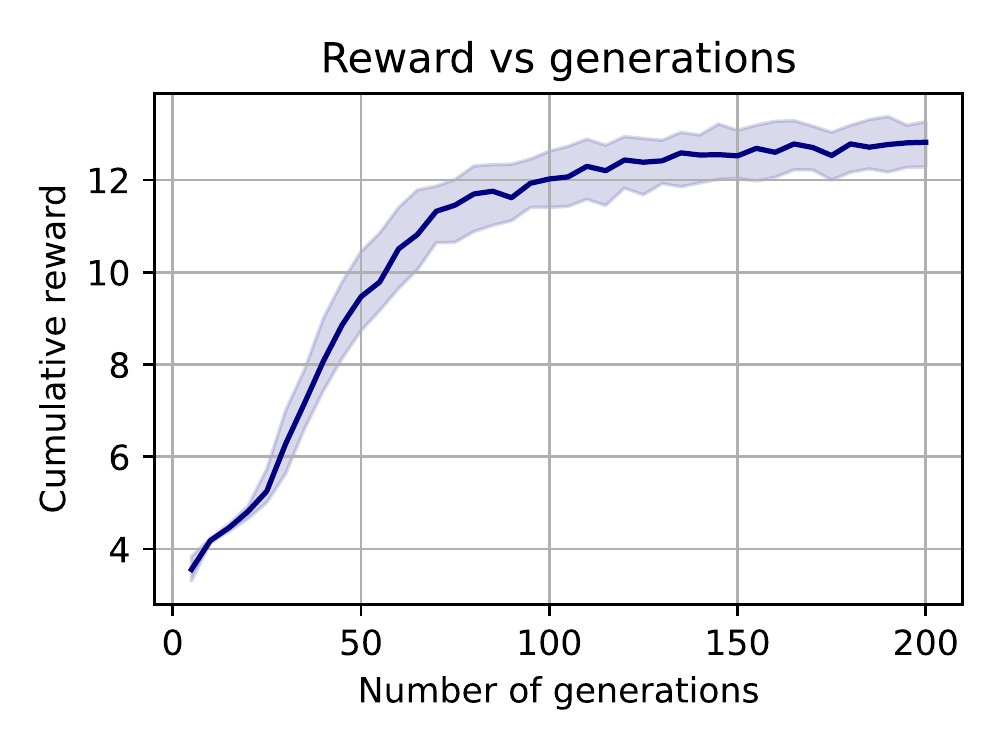}
            &
            \includegraphics[width=0.4\textwidth]{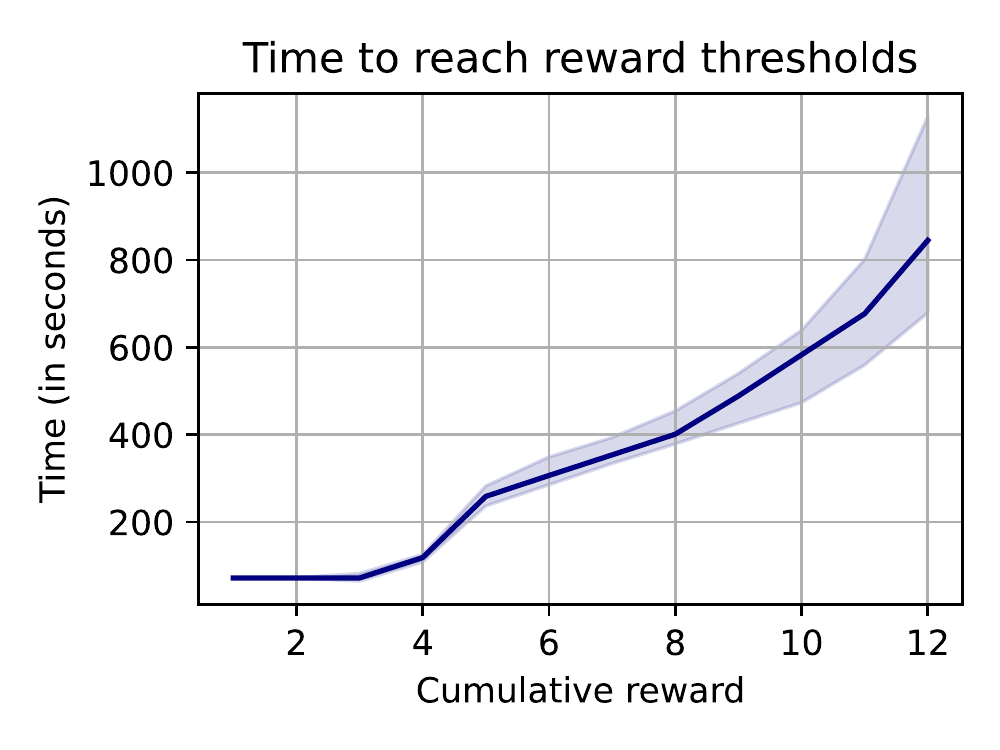}
        \end{tabular}
        \\
        { \footnotesize
        \underline{Policy}: A neural network with a single hidden recurrent layer of 64 neurons, with tanh activations.
        An elementwise non-affine layer normalization is applied just before the output layer.
        The output layer is a linear transformation, followed by a clip operation to make sure that the produced output is within valid boundaries.
        }
        \\
        \\
        \textbf{(b)} humanoid
        \\
        \begin{tabular}{c c}
            \includegraphics[width=0.4\textwidth]{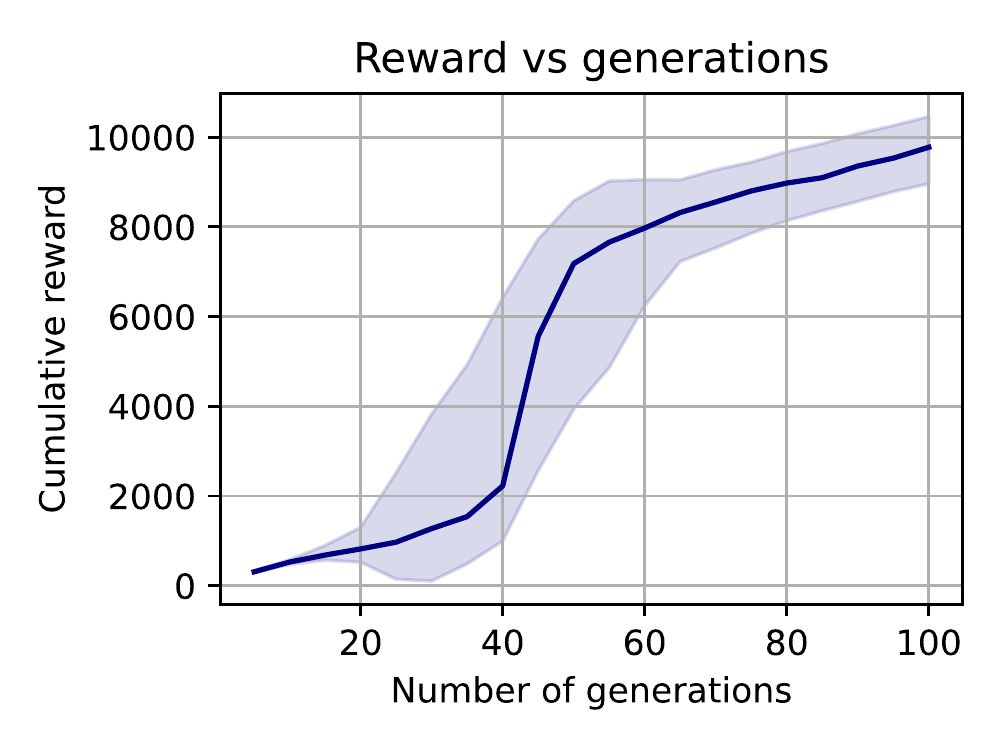}
            &
            \includegraphics[width=0.4\textwidth]{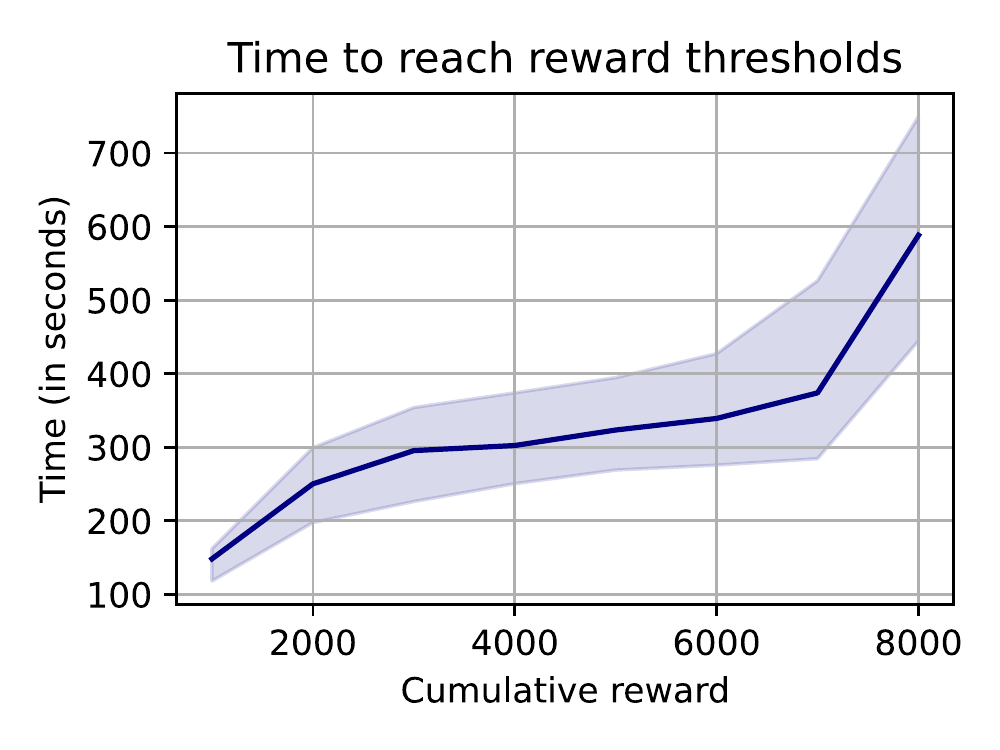}
        \end{tabular}
        \\
        { \footnotesize
        \underline{Policy}: A linear policy with bias terms.
        \underline{Alive bonus}: The default constant alive bonus of 5.0 (which has been reported to harm exploration for distribution-based evolutionary search \citep{mania2018,toklu2020clipup}) is removed. Instead, here we used a timestep-dependent alive bonus schedule:
        from timestep 0 to 400, the agent receives no alive bonus;
        from timestep 400 to 700, the alive bonus linearly increases from 0.0 to 10.0; and beginning with timestep 700, the agent receives an alive bonus of 10.0 for each timestep.
        By this tuning, we aim to make the alive bonus an active component of the reward function only towards the later phases of the evolution where the agents have learned to walk and their episodes are longer.
        }
        \\
        \\
        { \footnotesize
        \underline{PGPE+ClipUp configuration for both (a) and (b)}.
        We enabled two enhancements previously used by \citep{salimans2017}: observation normalization and 0-centered solution ranking (where the worst solution is assigned the rank -0.5 and the best solution is assigned the rank +0.5).
        Population size: 12000;
        center learning rate: 0.375;
        maximum speed for ClipUp: 0.75;
        standard deviation learning rate: 0.1.
        }
        \end{tabular}
    \caption{{\bf Results obtained on \texttt{brax} tasks with PGPE+ClipUp.} Each plot summarizes 30 runs. In each plot, bold curve represents the median and the shaded area is bounded by the mean $\pm$ the standard deviation.
    }
    \label{fig:brax}
\end{figure}

We performed test runs using PGPE+ClipUp on a single GPU for solving the \texttt{brax} tasks \emph{fetch} and \emph{humanoid}.
%
The results are shown in figure \ref{fig:brax}.
In the figure, we see that, in the median case, the cumulative reward threshold 12 can be reached in less than 14 minutes for the \emph{fetch} task.
In the case of the \emph{humanoid} task, if we accept 6000 as the solving threshold for the \emph{humanoid} task (as done for its MuJoCo counterpart, \emph{Humanoid-v4}, in e.g. \citep{salimans2017,mania2018}), we see that interesting locomotion behaviors are learned in less than 6 minutes in the median case.
We also see that a cumulative reward around 10000 can be achieved for the \emph{humanoid} task after 100 generations.

\subsection{Supervised learning}

There is a broad interest in applying evolutionary algorithms to supervised learning tasks, for example as an alternative to gradient-based methods for training neural networks for regression and classification \citep{mandischer2002comparison,zhang2017relationship,lehman2018more,lenc2019non,li2019ea}. Beyond simply serving as an alternative to gradient-descent, EA-based approaches to supervised learning open up a number of interesting research directions, such as training on inference-only hardware, optimizing for non-differentiable or discontinuous loss functions and optimizing non-differentiable neural architectures. To facilitate these research directions, \evotorch provides support for supervised neuroevolution out-of-the-box through the \texttt{SupervisedNE} problem class. 

To demonstrate this, we consider the experiments presented in \citep{lenc2019non}, specifically for training a small convolutional neural network on the MNIST dataset. We use `MNIST30K' network as described, except that we use layer normalization rather than batch normalization. With this network defined as a \texttt{PyTorch} module class \texttt{MNIST30K}, and with a training dataset \texttt{train\_dataset} prepared, creation of a problem instance is then straight-forward:

\begin{lstlisting}[language=Python]
    mnist_problem = SupervisedNE(
        train_dataset,  # Using the mnist training dataset
        MNIST30K,  # Training the MNIST30K module
        nn.CrossEntropyLoss(),  # Minimizing cross-entropy loss
        minibatch_size=1024,  # With a minibatch size of 512
        common_minibatch=True,  # Always using the same minibatch across all solutions on an actor
        num_actors=32,  # The total number of CPUs used
        num_gpus_per_actor='max',  # Dividing all available GPUs between the 32 actors
    )
\end{lstlisting}

In this configuration, 32 ray actors will be created, each with a fragment of a GPU assigned to it, for the forward (inference) step of evaluating each member of the population under cross-entropy loss across 1024 samples. As noted in \cite{lenc2019non}, gradient updates can be more stable when the same minibatch of data is used across all solutions evaluated by a single actor within a single step; this is reflected in the \texttt{common\_minibatch} argument. 

This problem class can be interfaced by search algorithms like any other. In this example, we use the PGPE search algorithm with the Adam optimizer and a population size of 3200. We generally found that raw fitnesses could be used directly, rather than rank-based fitness shaping, to beneficial effect. Additionally, as described by \cite{lenc2019non} as `semi-updates', the gradients used by distribution-based evolutionary algorithms can be approximated locally on individual actors and then averaged without introducing any bias; this is reflected in the \texttt{distributed} argument. Our search algorithm is therefore instantiated:

\begin{lstlisting}[language=Python]
    searcher = PGPE(
        mnist_problem, 
        radius_init=2.25,  # Initial radius of the search distribution 
        center_learning_rate=1e-2,  # Learning rate used by adam optimizer
        stdev_learning_rate=0.1,  # Learning rate for the standard deviation
        popsize=3200, 
        distributed=True,  # Gradients are computed locally at actors and averaged
        optimizer='adam',  # Using the adam optimizer
        ranking_method=None,  # No rank-based fitness shaping is used
    )
\end{lstlisting}

\begin{figure}[!t]
    \centering
    \includegraphics[width=0.5\textwidth]{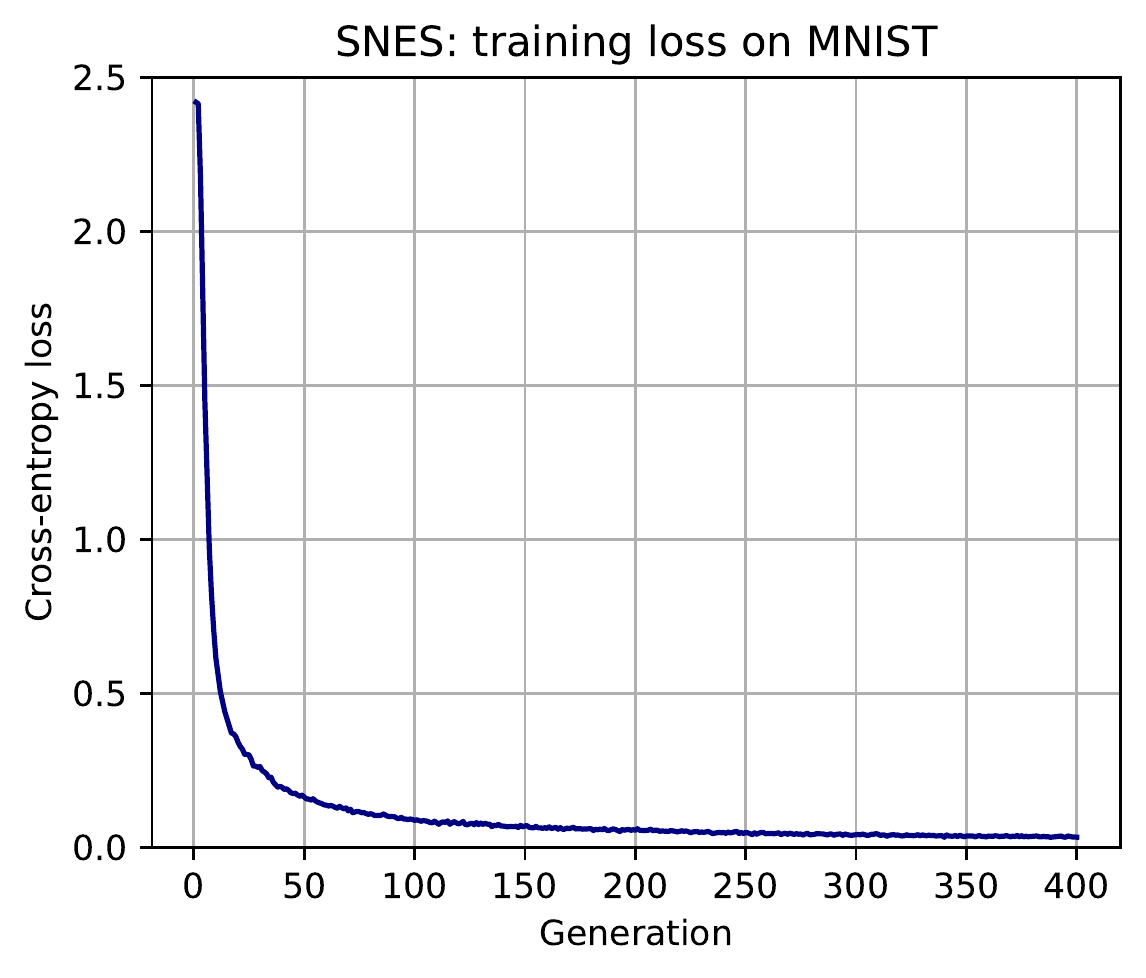}
    \caption{{\bf Training loss (cross entropy) of a single run of SNES on the MNIST dataset by generation.} The loss shown is the mean of the loss obtained by each member of the population in each generation.
    }
    \label{fig:training_mnist30k}
\end{figure}

Running the evolutionary algorithm for only $400$ generations, we observe quick convergence in the training loss as shown in Figure \ref{fig:training_mnist30k}. Once we have allowed the search algorithm to converge, we obtain a respectable $98.8\%$ test accuracy on unseen data. 

\subsection{Discrete Optimization}

In \evotorch, the data type of the decision variables can be declared as integer (e.g. \texttt{torch.int32}, \texttt{torch.int64}, etc.) or boolean (i.e. \texttt{torch.bool}). Assuming the goal of minimization, defining a discrete optimization problem looks like this:
\begin{lstlisting}[language=Python]
    discrete_problem = Problem("min", discrete_fitness_function, dtype=torch.int64, ...)
\end{lstlisting}
where \texttt{discrete\_fitness\_function} is expected as a function which receives a tensor whose \texttt{dtype} is \texttt{torch.int64} and returns a fitness tensor whose \texttt{dtype} is a floating-point type (\texttt{torch.float32} by default).

Such problems with discrete decision variable types can be solved with \texttt{GeneticAlgorithm}. \evotorch provides implementations of cross-over operators for \texttt{GeneticAlgorithm} that are friendly with discrete variable types. Problem-specific mutation operators can be defined by the user. A \texttt{GeneticAlgorithm} instantiation for \texttt{discrete\_problem} looks like this:
\begin{lstlisting}[language=Python]
    from evotorch.operators import TwoPointCrossOver

    def mutate(decision_values: torch.Tensor) -> torch.Tensor:
        # Here, decision_values is a 2-dimensional tensor where each row represents a solution.
        mutated_decision_values = ...  # a mutated copy of decision_values is generated here
        return mutated_decision_values

    ga = GeneticAlgorithm(
        discrete_problem,
        operators=[TwoPointCrossOver(discrete_problem), mutate],
        popsize=...,
        ...
    )
\end{lstlisting}
$ $

\noindent
\textbf{Challenges regarding vectorization.}
%
%
In many discrete optimization problems, straightforward vectorization might not be possible because the fitness functions must be expressed algorithmically (rather than as combinations of numeric PyTorch operations). An important example comes from the field of genetic programming where fitness functions interpret executable programs encoded in the candidate solutions. This interpreter must visit each symbol of the given program sequentially, perform conditional operations according to the symbol at hand, and produce the final result only at the end of the program. It is this algorithmic and sequential nature that hinders the vectorization across the decision variables of a batch of solutions.
However, this does \emph{not} mean that vectorization is hindered \emph{entirely}. Indeed, there can be a vectorized interpreter which, within a single step, processes the next symbol of \emph{each} program. Therefore, more generally, while algorithmic and/or sequential fitness functions might sacrifice vectorization across the decision variables, it can still be enjoyed across multiple solutions.

To allow the user to express sequential fitness functions in a practical manner, \evotorch provides vectorized PyTorch-based data structures that can be used as variable length lists, queues, stacks, and dictionaries. As an example, let us imagine that we have 10000 lists. Considering each of these lists, let us say we wish to sample a real number, and then, if the sampled number is positive, we wish to append it into its associated list. This program can be written without writing any loop as follows:

\begin{lstlisting}[language=Python]
    from evotorch.tools.structures import CList

    batch_of_lists = CList(  # make a batch of lists with contiguous storage
        max_length=10,  # each list can have at most this many elements
        batch_size=10000,   # we want this many lists in our batch
        # device="cuda",  # can be enabled for exploiting the GPU
    )

    batch_of_numbers = torch.randn(10000, device=batch_of_lists.device)
    batch_of_lists.append_(batch_of_numbers, where=(batch_of_numbers > 0))
\end{lstlisting}

\begin{figure}[!t]
    \centering
    \includegraphics[width=0.5\textwidth]{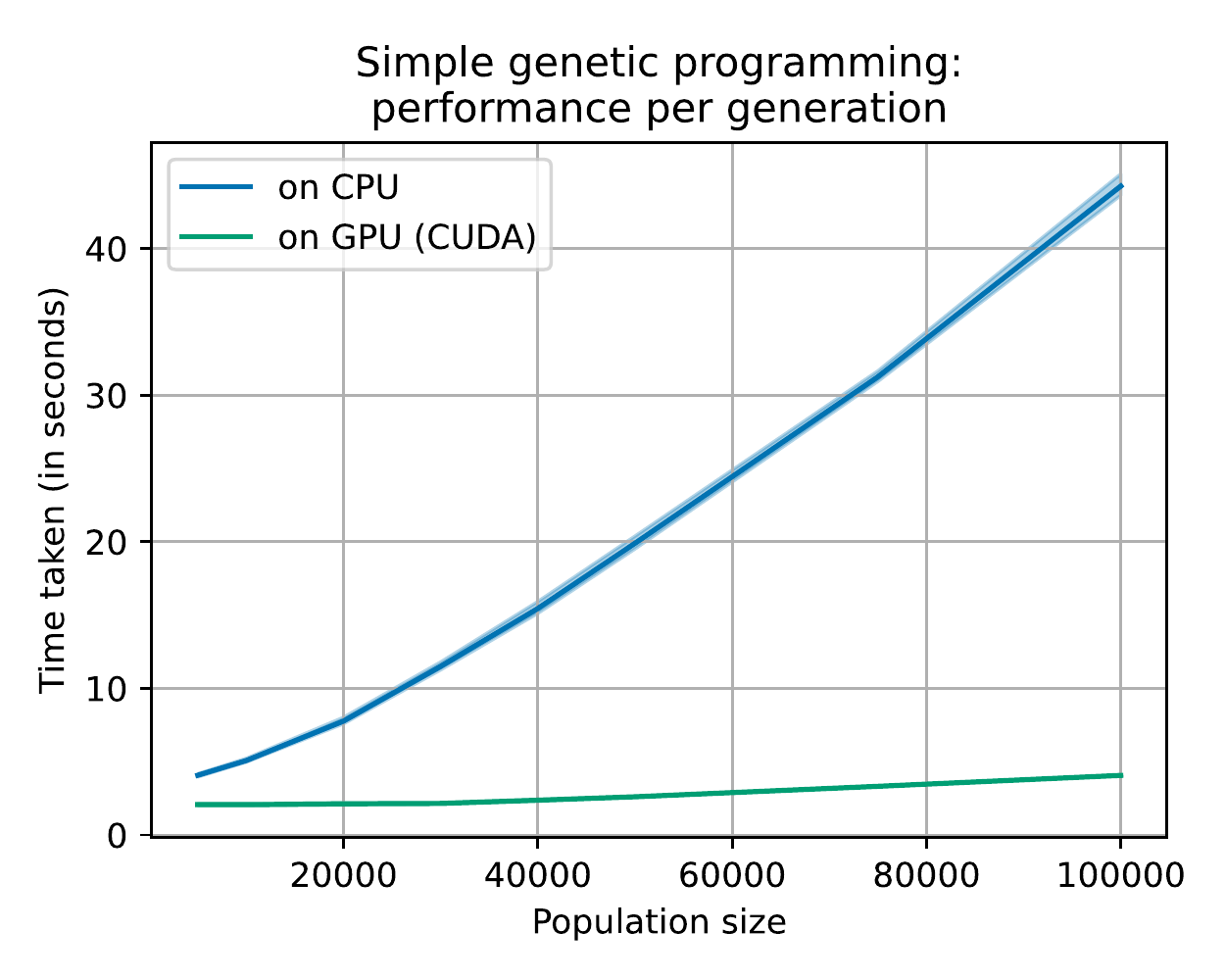}
    \caption{{\bf GPU vs CPU wall-clock times for a simple genetic programming notebook per generation.}  Bold curves represent the medians, and the shaded regions represent the means $\pm$ the standard deviations.
    For each population size, 30 runs were started, each run lasting for 50 generations. Median time was summarized over all generations (except the very first generation where the population initialization procedures have an overhead) of all runs. Therefore, each reported time for each population size is a summary over 49 $\times$ 30 = 1470 data points.
    }
    \label{fig:gpresults}
\end{figure}

A genetic programming example has been implemented where the fitness function is a stack-based interpreter written using the \texttt{CList} structure.  Populations ranged from 5000 to 100000 programs of maximum length 20.  Wall-clock times were compared for a single GPU versus a 40-thread CPU (without multiple actors, with multithreading enabled for PyTorch).
The results in figure~\ref{fig:gpresults} show that the 
GPU time requirement is almost flat up to population size 30000, whereas the CPU curve increases sharply, taking more than 10 times that the GPU implementation for population size 10000.

Although we have used genetic programming as our example, we argue that fitness functions of similar nature might be encountered in other discrete optimization fields such as transportation, scheduling, etc. It is our hope that the features of \evotorch mentioned here can allow the users to quickly prototype GPU-friendly problem definitions in such cases.

\section{Conclusion}
We have presented \evotorch, an evolutionary computation library written in Python programming language and built upon \texttt{PyTorch} and \texttt{ray}.
Among the highlights of the library are:

\begin{itemize}
    \item seamless interaction with the well-developed scientific ecosystem of Python;
    \item ability to exploit hardware accelerators such as GPU with the help of \texttt{PyTorch};
    \item ability to scale up to large experiments with the help of CPU-based parallelization and distributed computation abilities brought by \texttt{ray};
    \item efficient neuroevolution and blackbox optimization thanks to having state-of-the-art algorithm implementations (PGPE, XNES, etc.);
    \item general abilities to address various types of problems (with single or multiple objectives, with continuous or discrete variable types, etc.).
\end{itemize}

It is our hope that \evotorch will become a practical tool for performing large-scale evolutionary computation in both the research and industrial communities.

\section*{Acknowledgements}
We are grateful for the feedback we have received \citep{evotorchissues} and for the contributed code \citep{evotorchcontributors}.
We thank Faustino Gomez for his comments and for reviewing the paper.

\bibliographystyle{humannat}
\bibliography{evotorch}

\begin{thebibliography}{}

\bibitem[\protect\astroncite{Abadi et~al.}{2016}]{tensorflow2016}
Abadi, M., P.~Barham, J.~Chen, Z.~Chen, A.~Davis, J.~Dean, M.~Devin,
  S.~Ghemawat, G.~Irving, M.~Isard, M.~Kudlur, J.~Levenberg, R.~Monga,
  S.~Moore, D.~G. Murray, B.~Steiner, P.~Tucker, V.~Vasudevan, P.~Warden,
  M.~Wicke, Y.~Yu, and X.~Zheng\leavevmode\nopagebreak\newline 2016.
\newblock {TensorFlow}: {A} system for large-scale machine learning.
\newblock In {\em 12th USENIX Symposium on Operating Systems Design and
  Implementation (OSDI 16)}, Pp.~ 265--283.
\newblock
  \online{https://www.usenix.org/system/files/conference/osdi16/osdi16-abadi.pdf}.

\bibitem[\protect\astroncite{Back and Schwefel}{1996}]{back1996evolutionary}
Back, T. and H.-P. Schwefel\leavevmode\nopagebreak\newline 1996.
\newblock Evolutionary computation: An overview.
\newblock In {\em Proceedings of IEEE International Conference on Evolutionary
  Computation}, Pp.~ 20--29. IEEE.
\newblock \doi{10.1109/ICEC.1996.542329}.

\bibitem[\protect\astroncite{Biscani and Izzo}{2020}]{Biscani2020}
Biscani, F. and D.~Izzo\leavevmode\nopagebreak\newline 2020.
\newblock A parallel global multiobjective framework for optimization: pagmo.
\newblock {\em Journal of Open Source Software}, 5(53):2338.

\bibitem[\protect\astroncite{Blank and Deb}{2020}]{pymoo2020}
Blank, J. and K.~Deb\leavevmode\nopagebreak\newline 2020.
\newblock Pymoo: Multi-objective optimization in {P}ython.
\newblock {\em IEEE Access}, 8:89497--89509.
\newblock \doi{10.1109/ACCESS.2020.2990567}.

\bibitem[\protect\astroncite{Brockman et~al.}{2016}]{brockman2016openai}
Brockman, G., V.~Cheung, L.~Pettersson, J.~Schneider, J.~Schulman, J.~Tang, and
  W.~Zaremba\leavevmode\nopagebreak\newline 2016.
\newblock Openai gym.
\newblock {\em arXiv preprint}.
\newblock \arxiv{1606.01540}.

\bibitem[\protect\astroncite{Chellapilla et~al.}{2006}]{chellapilla2006}
Chellapilla, K., S.~Puri, and P.~Simard\leavevmode\nopagebreak\newline 2006.
\newblock High performance convolutional neural networks for document
  processing.
\newblock In {\em Tenth international workshop on frontiers in handwriting
  recognition (IWFHR 10)}.
\newblock \hal{inria-00112631}.

\bibitem[\protect\astroncite{Ciresan et~al.}{2011}]{ciresan2011}
Ciresan, D.~C., U.~Meier, J.~Masci, L.~M. Gambardella, and
  J.~Schmidhuber\leavevmode\nopagebreak\newline 2011.
\newblock Flexible, high performance convolutional neural networks for image
  classification.
\newblock In {\em Proceedings of the Twenty-Second International Joint
  Conference on Artificial Intelligence (IJCAI'11)}, Pp.~ 1237--1242. AAAI
  Press.
\newblock
  \online{https://www.aaai.org/ocs/index.php/IJCAI/IJCAI11/paper/view/3098/3425}.

\bibitem[\protect\astroncite{Coumans and Bai}{2021}]{coumans2021}
Coumans, E. and Y.~Bai\leavevmode\nopagebreak\newline 2016--2021.
\newblock Pybullet, a python module for physics simulation for games, robotics
  and machine learning.
\newblock \url{http://pybullet.org}.

\bibitem[\protect\astroncite{Deb and Jain}{2013}]{deb2013evolutionary}
Deb, K. and H.~Jain\leavevmode\nopagebreak\newline 2013.
\newblock An evolutionary many-objective optimization algorithm using
  reference-point-based nondominated sorting approach, part {I}: solving
  problems with box constraints.
\newblock {\em IEEE Transactions on Evolutionary Computation (TEVC)},
  18(4):577--601.
\newblock \doi{10.1109/TEVC.2013.2281535}.

\bibitem[\protect\astroncite{Deb et~al.}{2002}]{deb2002}
Deb, K., A.~Pratap, S.~Agarwal, and T.~Meyarivan\leavevmode\nopagebreak\newline
  2002.
\newblock A fast and elitist multiobjective genetic algorithm: {NSGA}-{II}.
\newblock {\em IEEE Transactions on Evolutionary Computation (TEVC)},
  6(2):182--197.
\newblock \doi{10.1109/4235.996017}.

\bibitem[\protect\astroncite{Ellenberger}{2019}]{benelot2018}
Ellenberger, B.\leavevmode\nopagebreak\newline 2018--2019.
\newblock Pybullet gymperium.
\newblock \url{ https://github.com/benelot/pybullet-gym}.

\bibitem[\protect\astroncite{Erez et~al.}{2011}]{erez2011}
Erez, T., Y.~Tassa, and E.~Todorov\leavevmode\nopagebreak\newline 2011.
\newblock Infinite-horizon model predictive control for periodic tasks with
  contacts.
\newblock In {\em Proceedings of Robotics: Science and Systems VII}. MIT Press.
\newblock \doi{10.15607/RSS.2011.VII.010}.

\bibitem[\protect\astroncite{Fey and Lenssen}{2019}]{fey2019fast}
Fey, M. and J.~E. Lenssen\leavevmode\nopagebreak\newline 2019.
\newblock Fast graph representation learning with pytorch geometric.
\newblock {\em arXiv preprint}.
\newblock \arxiv{1903.02428}.

\bibitem[\protect\astroncite{Fortin et~al.}{2012}]{deap2012}
Fortin, F.-A., F.-M. De~Rainville, M.-A.~G. Gardner, M.~Parizeau, and
  C.~Gagn{\'e}\leavevmode\nopagebreak\newline 2012.
\newblock {DEAP}: Evolutionary algorithms made easy.
\newblock {\em Journal of Machine Learning Research (JMLR)}, 13(1):2171--2175.
\newblock \online{https://www.jmlr.org/papers/v13/fortin12a.html}.

\bibitem[\protect\astroncite{Freeman et~al.}{2021a}]{brax2021github}
Freeman, C.~D., E.~Frey, A.~Raichuk, S.~Girgin, I.~Mordatch, and
  O.~Bachem\leavevmode\nopagebreak\newline 2021a.
\newblock Brax - a differentiable physics engine for large scale rigid body
  simulation.
\newblock \url{http://github.com/google/brax}.

\bibitem[\protect\astroncite{Freeman et~al.}{2021b}]{freeman2021brax}
Freeman, C.~D., E.~Frey, A.~Raichuk, S.~Girgin, I.~Mordatch, and
  O.~Bachem\leavevmode\nopagebreak\newline 2021b.
\newblock Brax--a differentiable physics engine for large scale rigid body
  simulation.
\newblock {\em arXiv preprint}.
\newblock \arxiv{2106.13281}.

\bibitem[\protect\astroncite{Frostig et~al.}{2018}]{frostig2018}
Frostig, R., M.~J. Johnson, and C.~Leary\leavevmode\nopagebreak\newline 2018.
\newblock Compiling machine learning programs via high-level tracing.
\newblock {\em Systems for Machine Learning}, 4(9).
\newblock \online{https://mlsys.org/Conferences/doc/2018/146.pdf}.

\bibitem[\protect\astroncite{Gardner et~al.}{2017}]{Gardner2017AllenNLP}
Gardner, M., J.~Grus, M.~Neumann, O.~Tafjord, P.~Dasigi, N.~F. Liu, M.~Peters,
  M.~Schmitz, and L.~S. Zettlemoyer\leavevmode\nopagebreak\newline 2017.
\newblock Allennlp: A deep semantic natural language processing platform.
\newblock {\em arXiv preprint}.
\newblock \arxiv{1803.07640}.

\bibitem[\protect\astroncite{{GitHub}}{2023a}]{evotorchcontributors}
{GitHub}\leavevmode\nopagebreak\newline 2023a.
\newblock Contributors to nnaisense/evotorch.
\newblock \url{https://github.com/nnaisense/evotorch/graphs/contributors}.

\bibitem[\protect\astroncite{{GitHub}}{2023b}]{evotorchissues}
{GitHub}\leavevmode\nopagebreak\newline 2023b.
\newblock Issues $\cdot$ nnaisense/evotorch.
\newblock \url{https://github.com/nnaisense/evotorch/issues}.

\bibitem[\protect\astroncite{Glasmachers et~al.}{2010}]{glasmachers2010}
Glasmachers, T., T.~Schaul, S.~Yi, D.~Wierstra, and
  J.~Schmidhuber\leavevmode\nopagebreak\newline 2010.
\newblock Exponential natural evolution strategies.
\newblock In {\em Proceedings of the 12th annual conference on Genetic and
  Evolutionary Computation (GECCO'10)}, Pp.~ 393--400. ACM.
\newblock \doi{10.1145/1830483.1830557}.

\bibitem[\protect\astroncite{Gomez et~al.}{2008}]{gomez2008accelerated}
Gomez, F., J.~Schmidhuber, R.~Miikkulainen, and
  M.~Mitchell\leavevmode\nopagebreak\newline 2008.
\newblock Accelerated neural evolution through cooperatively coevolved
  synapses.
\newblock {\em Journal of Machine Learning Research}, 9(5).
\newblock \online{https://www.jmlr.org/papers/v9/gomez08a.html}.

\bibitem[\protect\astroncite{Greff et~al.}{2017}]{greff2017}
Greff, K., A.~Klein, M.~Chovanec, F.~Hutter, and
  J.~Schmidhuber\leavevmode\nopagebreak\newline 2017.
\newblock {T}he {S}acred {I}nfrastructure for {C}omputational {R}esearch.
\newblock In {\em Proceedings of the 16th {P}ython in science conference
  (Sci{P}y 2017)}, Pp.~ 49--56.
\newblock \doi{10.25080/shinma-7f4c6e7-008}.

\bibitem[\protect\astroncite{Ha}{2019}]{ha2019}
Ha, D.\leavevmode\nopagebreak\newline 2019.
\newblock Reinforcement learning for improving agent design.
\newblock {\em Artificial Life}, 25(4):352--365.
\newblock \doi{10.1162/artl\_a\_00301}.

\bibitem[\protect\astroncite{Hansen et~al.}{2010}]{hansen2010comparing}
Hansen, N., A.~Auger, R.~Ros, S.~Finck, and
  P.~Po{\v{s}}{\'\i}k\leavevmode\nopagebreak\newline 2010.
\newblock Comparing results of 31 algorithms from the black-box optimization
  benchmarking {BBOB}-2009.
\newblock In {\em Proceedings of the 12th annual conference on Genetic and
  Evolutionary Computation (GECCO'10)}, Pp.~ 1689--1696. ACM.
\newblock \doi{10.1145/1830761.1830790}.

\bibitem[\protect\astroncite{Hansen and Ostermeier}{2001}]{hansen2001}
Hansen, N. and A.~Ostermeier\leavevmode\nopagebreak\newline 2001.
\newblock Completely derandomized self-adaptation in evolution strategies.
\newblock {\em Evolutionary computation}, 9(2):159--195.
\newblock \doi{10.1162/106365601750190398}.

\bibitem[\protect\astroncite{Hoffmeister and
  B{\"a}ck}{1990}]{hoffmeister1990genetic}
Hoffmeister, F. and T.~B{\"a}ck\leavevmode\nopagebreak\newline 1990.
\newblock Genetic algorithms and evolution strategies: Similarities and
  differences.
\newblock In {\em International Conference on Parallel Problem Solving from
  Nature (PPSN 1990)}, Pp.~ 455--469. Springer.
\newblock \doi{10.1007/bfb0029787}.

\bibitem[\protect\astroncite{Holland}{1975}]{holland1975}
Holland, J.\leavevmode\nopagebreak\newline 1975.
\newblock {\em Adaptation in Natural and Artificial Systems}.
\newblock University of Michigan Press.
\newblock \doi{10.7551/mitpress/1090.001.0001}.

\bibitem[\protect\astroncite{Holland}{1992}]{holland1992}
Holland, J.~H.\leavevmode\nopagebreak\newline 1992.
\newblock Genetic algorithms.
\newblock {\em Scientific american}, 267(1):66--73.
\newblock \online{https://www.jstor.org/stable/24939139}.

\bibitem[\protect\astroncite{Huang et~al.}{2023}]{huang2023}
Huang, B., R.~Cheng, Y.~Jin, and K.~C. Tan\leavevmode\nopagebreak\newline 2023.
\newblock {EvoX}: A distributed {GPU}-accelerated library towards scalable
  evolutionary computation.
\newblock {\em arXiv preprint}.
\newblock \arxiv{2301.12457}.

\bibitem[\protect\astroncite{Innes}{2018}]{innes2018}
Innes, M.\leavevmode\nopagebreak\newline 2018.
\newblock Flux: {Elegant} machine learning with {Julia}.
\newblock {\em Journal of Open Source Software}, 3(25):602.
\newblock \doi{10.21105/joss.00602}.

\bibitem[\protect\astroncite{{J}ames {B}ergstra et~al.}{2010}]{theano2010}
{J}ames {B}ergstra, {O}livier {B}reuleux, {F}r\'ed\'eric {B}astien, {P}ascal
  {L}amblin, {R}azvan {P}ascanu, {G}uillaume {D}esjardins, {J}oseph {T}urian,
  {D}avid~{W}arde {F}arley, and {Y}oshua
  {B}engio\leavevmode\nopagebreak\newline 2010.
\newblock {T}heano: a cpu and gpu math compiler in {P}ython.
\newblock In {\em Proceedings of the 9th {P}ython in science conference
  (Sci{P}y 2010)}, Pp.~ 18 -- 24.
\newblock \doi{10.25080/Majora-92bf1922-003}.

\bibitem[\protect\astroncite{Janson et~al.}{2008}]{janson2008molecular}
Janson, S., D.~Merkle, and M.~Middendorf\leavevmode\nopagebreak\newline 2008.
\newblock Molecular docking with multi-objective particle swarm optimization.
\newblock {\em Applied Soft Computing}, 8(1):666--675.
\newblock \doi{10.1016/j.asoc.2007.05.005}.

\bibitem[\protect\astroncite{Kingma and Ba}{2015}]{kingma2015}
Kingma, D.~P. and J.~Ba\leavevmode\nopagebreak\newline 2015.
\newblock Adam: A method for stochastic optimization.
\newblock In {\em Proceedings of 3rd International Conference on Learning
  Representations (ICLR)}.
\newblock \arxiv{1412.6980}.

\bibitem[\protect\astroncite{Klimov and Schulman}{2017}]{klimov2017}
Klimov, O. and J.~Schulman\leavevmode\nopagebreak\newline 2017.
\newblock Roboschool.
\newblock \url{https://openai.com/blog/roboschool/}.

\bibitem[\protect\astroncite{Krizhevsky et~al.}{2012}]{krizhevsky2012}
Krizhevsky, A., I.~Sutskever, and G.~E. Hinton\leavevmode\nopagebreak\newline
  2012.
\newblock {ImageNet} classification with deep convolutional neural networks.
\newblock {\em Advances in neural information processing systems (NIPS 2012)},
  25.
\newblock
  \online{https://proceedings.neurips.cc/paper/2012/file/c399862d3b9d6b76c8436e924a68c45b-Paper.pdf}.

\bibitem[\protect\astroncite{Kursawe}{1990}]{kursawe1991variant}
Kursawe, F.\leavevmode\nopagebreak\newline 1990.
\newblock A variant of evolution strategies for vector optimization.
\newblock In {\em International conference on parallel problem solving from
  nature (PPSN 1990)}, Pp.~ 193--197. Springer.
\newblock \doi{10.1007/BFb0029752}.

\bibitem[\protect\astroncite{Lange}{2022}]{evosax2022}
Lange, R.~T.\leavevmode\nopagebreak\newline 2022.
\newblock evosax: Jax-based evolution strategies.
\newblock {\em arXiv preprint}.
\newblock \arxiv{2212.04180}.

\bibitem[\protect\astroncite{Lehman et~al.}{2018}]{lehman2018more}
Lehman, J., J.~Chen, J.~Clune, and K.~O. Stanley\leavevmode\nopagebreak\newline
  2018.
\newblock {ES} is more than just a traditional finite-difference approximator.
\newblock In {\em Proceedings of the 16th annual conference on Genetic and
  Evolutionary Computation (GECCO'18)}, Pp.~ 450--457.
\newblock \doi{10.1145/3205455.3205474}.

\bibitem[\protect\astroncite{Lenc et~al.}{2019}]{lenc2019non}
Lenc, K., E.~Elsen, T.~Schaul, and K.~Simonyan\leavevmode\nopagebreak\newline
  2019.
\newblock Non-differentiable supervised learning with evolution strategies and
  hybrid methods.
\newblock {\em arXiv preprint}.
\newblock \arxiv{1906.03139}.

\bibitem[\protect\astroncite{Li et~al.}{2019}]{li2019ea}
Li, Y., Z.~Zhu, D.~Kong, H.~Han, and Y.~Zhao\leavevmode\nopagebreak\newline
  2019.
\newblock {EA-LSTM}: Evolutionary attention-based {LSTM} for time series
  prediction.
\newblock {\em Knowledge-Based Systems}, 181:104785.
\newblock \doi{10.1016/j.knosys.2019.05.028}.

\bibitem[\protect\astroncite{Lu et~al.}{2019}]{lu2019nsga}
Lu, Z., I.~Whalen, V.~Boddeti, Y.~Dhebar, K.~Deb, E.~Goodman, and
  W.~Banzhaf\leavevmode\nopagebreak\newline 2019.
\newblock {NSGA}-net: neural architecture search using multi-objective genetic
  algorithm.
\newblock In {\em Proceedings of the 17th annual conference on Genetic and
  Evolutionary Computation (GECCO'19)}, Pp.~ 419--427.
\newblock \doi{10.1145/3321707.3321729}.

\bibitem[\protect\astroncite{Luke}{1998}]{Luke1998ECJSoftware}
Luke, S.\leavevmode\nopagebreak\newline 1998.
\newblock {ECJ} evolutionary computation library.
\newblock Available for free at \url{http://cs.gmu.edu/~eclab/projects/ecj/}.

\bibitem[\protect\astroncite{Mandischer}{2002}]{mandischer2002comparison}
Mandischer, M.\leavevmode\nopagebreak\newline 2002.
\newblock A comparison of evolution strategies and backpropagation for neural
  network training.
\newblock {\em Neurocomputing}, 42(1-4):87--117.
\newblock \doi{10.1016/S0925-2312(01)00596-3}.

\bibitem[\protect\astroncite{Mania et~al.}{2018}]{mania2018}
Mania, H., A.~Guy, and B.~Recht\leavevmode\nopagebreak\newline 2018.
\newblock Simple random search of static linear policies is competitive for
  reinforcement learning.
\newblock In {\em Advances in Neural Information Processing Systems (NeurIPS
  2018)}, Pp.~ 1800--1809.
\newblock
  \online{https://proceedings.neurips.cc/paper/2018/file/7634ea65a4e6d9041cfd3f7de18e334a-Paper.pdf}.

\bibitem[\protect\astroncite{{MLflow Contributors}}{2022}]{mlflow}
{MLflow Contributors}\leavevmode\nopagebreak\newline 2022.
\newblock mlflow: Open source platform for the machine learning lifecycle.
\newblock \url{https://github.com/mlflow/mlflow}.

\bibitem[\protect\astroncite{Moritz et~al.}{2018}]{ray2018}
Moritz, P., R.~Nishihara, S.~Wang, A.~Tumanov, R.~Liaw, E.~Liang, M.~Elibol,
  Z.~Yang, W.~Paul, M.~I. Jordan, and I.~Stoica\leavevmode\nopagebreak\newline
  2018.
\newblock Ray: A distributed framework for emerging {AI} applications.
\newblock In {\em 13th USENIX Symposium on Operating Systems Design and
  Implementation (OSDI 18)}, Pp.~ 561--577.

\bibitem[\protect\astroncite{Mouret and Clune}{2015}]{mouret2015illuminating}
Mouret, J.-B. and J.~Clune\leavevmode\nopagebreak\newline 2015.
\newblock Illuminating search spaces by mapping elites.
\newblock {\em arXiv preprint}.
\newblock \arxiv{1504.04909}.

\bibitem[\protect\astroncite{{Neptune Contributors}}{2022}]{neptune}
{Neptune Contributors}\leavevmode\nopagebreak\newline 2022.
\newblock neptune-client: Experiment tracking tool and model registry.
\newblock \url{https://github.com/neptune-ai/neptune-client}.

\bibitem[\protect\astroncite{Panichella}{2019}]{panichella2019adaptive}
Panichella, A.\leavevmode\nopagebreak\newline 2019.
\newblock An adaptive evolutionary algorithm based on non-euclidean geometry
  for many-objective optimization.
\newblock In {\em Proceedings of the 17th annual conference on Genetic and
  Evolutionary Computation (GECCO'19)}, Pp.~ 595--603.
\newblock \doi{10.1145/3321707.3321839}.

\bibitem[\protect\astroncite{Paszke et~al.}{2019}]{pytorch2019}
Paszke, A., S.~Gross, F.~Massa, A.~Lerer, J.~Bradbury, G.~Chanan, T.~Killeen,
  Z.~Lin, N.~Gimelshein, L.~Antiga, A.~Desmaison, A.~Kopf, E.~Yang, Z.~DeVito,
  M.~Raison, A.~Tejani, S.~Chilamkurthy, B.~Steiner, L.~Fang, J.~Bai, and
  S.~Chintala\leavevmode\nopagebreak\newline 2019.
\newblock {PyTorch}: An imperative style, high-performance deep learning
  library.
\newblock In {\em Advances in Neural Information Processing Systems (NeurIPS
  2019)}, volume~32.
\newblock
  \online{https://proceedings.neurips.cc/paper/2019/file/bdbca288fee7f92f2bfa9f7012727740-Paper.pdf}.

\bibitem[\protect\astroncite{Rubinstein}{1999}]{rubinstein1999}
Rubinstein, R.\leavevmode\nopagebreak\newline 1999.
\newblock The cross-entropy method for combinatorial and continuous
  optimization.
\newblock {\em Methodology and computing in applied probability},
  1(2):127--190.
\newblock \doi{10.1023/A:1010091220143}.

\bibitem[\protect\astroncite{Rubinstein}{1997}]{rubinstein1997}
Rubinstein, R.~Y.\leavevmode\nopagebreak\newline 1997.
\newblock Optimization of computer simulation models with rare events.
\newblock {\em European Journal of Operational Research}, 99(1):89--112.
\newblock \doi{10.1016/S0377-2217(96)00385-2}.

\bibitem[\protect\astroncite{Salimans et~al.}{2017}]{salimans2017}
Salimans, T., J.~Ho, X.~Chen, S.~Sidor, and
  I.~Sutskever\leavevmode\nopagebreak\newline 2017.
\newblock Evolution strategies as a scalable alternative to reinforcement
  learning.
\newblock {\em arXiv preprint arXiv}.
\newblock \arxiv{1703.03864}.

\bibitem[\protect\astroncite{Sarker and Ray}{2009}]{sarker2009improved}
Sarker, R. and T.~Ray\leavevmode\nopagebreak\newline 2009.
\newblock An improved evolutionary algorithm for solving multi-objective crop
  planning models.
\newblock {\em Computers and electronics in agriculture}, 68(2):191--199.
\newblock \doi{10.1016/j.compag.2009.06.002}.

\bibitem[\protect\astroncite{Schaul et~al.}{2011}]{schaul2011}
Schaul, T., T.~Glasmachers, and J.~Schmidhuber\leavevmode\nopagebreak\newline
  2011.
\newblock High dimensions and heavy tails for natural evolution strategies.
\newblock In {\em Proceedings of the 13th annual conference on Genetic and
  Evolutionary Computation (GECCO'11)}, Pp.~ 845--852.
\newblock \doi{10.1145/2001576.2001692}.

\bibitem[\protect\astroncite{Schmidhuber}{2015}]{schmidhuber2015}
Schmidhuber, J.\leavevmode\nopagebreak\newline 2015.
\newblock Deep learning in neural networks: An overview.
\newblock {\em Neural networks}, 61:85--117.
\newblock \doi{10.1016/j.neunet.2014.09.003}.

\bibitem[\protect\astroncite{Sehnke et~al.}{2010}]{sehnke2010}
Sehnke, F., C.~Osendorfer, T.~R{\"u}ckstie{\ss}, A.~Graves, J.~Peters, and
  J.~Schmidhuber\leavevmode\nopagebreak\newline 2010.
\newblock Parameter-exploring policy gradients.
\newblock {\em Neural Networks}, 23(4):551--559.
\newblock \doi{10.1016/j.neunet.2009.12.004}.

\bibitem[\protect\astroncite{Tang et~al.}{2022}]{evojax2022}
Tang, Y., Y.~Tian, and D.~Ha\leavevmode\nopagebreak\newline 2022.
\newblock {EvoJAX}: Hardware-accelerated neuroevolution.
\newblock {\em arXiv preprint}.
\newblock \arxiv{2202.05008}.

\bibitem[\protect\astroncite{Tassa et~al.}{2012}]{tassa2012}
Tassa, Y., T.~Erez, and E.~Todorov\leavevmode\nopagebreak\newline 2012.
\newblock Synthesis and stabilization of complex behaviors through online
  trajectory optimization.
\newblock In {\em Proceedings of the IEEE/RSJ International Conference on
  Intelligent Robots and Systems}, Pp.~ 4906--4913. IEEE.
\newblock \doi{10.1109/IROS.2012.6386025}.

\bibitem[\protect\astroncite{Todorov et~al.}{2012}]{todorov2012mujoco}
Todorov, E., T.~Erez, and Y.~Tassa\leavevmode\nopagebreak\newline 2012.
\newblock Mujoco: A physics engine for model-based control.
\newblock In {\em Proceedings of the IEEE/RSJ International Conference on
  Intelligent Robots and Systems}, Pp.~ 5026--5033. IEEE.
\newblock \doi{10.1109/IROS.2012.6386109}.

\bibitem[\protect\astroncite{Toklu et~al.}{2020}]{toklu2020clipup}
Toklu, N.~E., P.~Liskowski, and R.~K. Srivastava\leavevmode\nopagebreak\newline
  2020.
\newblock Clipup: a simple and powerful optimizer for distribution-based policy
  evolution.
\newblock In {\em International Conference on Parallel Problem Solving from
  Nature (PPSN 2020)}, Pp.~ 515--527. Springer.
\newblock \doi{10.1007/978-3-030-58115-2\_36}.

\bibitem[\protect\astroncite{Tokui et~al.}{2015}]{chainer2015}
Tokui, S., K.~Oono, S.~Hido, and J.~Clayton\leavevmode\nopagebreak\newline
  2015.
\newblock Chainer: a next-generation open source framework for deep learning.
\newblock In {\em Proceedings of workshop on Machine Learning Systems
  (LearningSys) in Advances in Neural information processing systems (NIPS
  2015)}, volume~5, Pp.~ 1--6.
\newblock
  \online{http://learningsys.org/papers/LearningSys\_2015\_paper\_33.pdf}.

\bibitem[\protect\astroncite{{Van Rossum} and Drake}{2009}]{vanrossum2009}
{Van Rossum}, G. and F.~L. Drake\leavevmode\nopagebreak\newline 2009.
\newblock {\em Python 3 Reference Manual}.
\newblock CreateSpace.
\newblock \online{https://docs.python.org/3/reference/}.

\bibitem[\protect\astroncite{Wu et~al.}{2019}]{wu2019detectron2}
Wu, Y., A.~Kirillov, F.~Massa, W.-Y. Lo, and
  R.~Girshick\leavevmode\nopagebreak\newline 2019.
\newblock Detectron2.
\newblock \url{https://github.com/facebookresearch/detectron2}.

\bibitem[\protect\astroncite{Yuret}{2016}]{yuret2016}
Yuret, D.\leavevmode\nopagebreak\newline 2016.
\newblock Knet: {Beginning} deep learning with 100 lines of {Julia}.
\newblock In {\em Proceedings of workshop on Machine Learning Systems
  (LearningSys) in Advances in Neural information processing systems (NIPS
  2016)}, volume 2016, P.~~5.
\newblock
  \online{https://www.vliz.be/nl/kaarten-bibliotheek?module=ref\&refid=310930}.

\bibitem[\protect\astroncite{Zhang and Li}{2007}]{zhang2007moea}
Zhang, Q. and H.~Li\leavevmode\nopagebreak\newline 2007.
\newblock {MOEA/D}: A multiobjective evolutionary algorithm based on
  decomposition.
\newblock {\em IEEE Transactions on evolutionary computation (TEVC)},
  11(6):712--731.
\newblock \doi{10.1109/TEVC.2007.892759}.

\bibitem[\protect\astroncite{Zhang et~al.}{2017}]{zhang2017relationship}
Zhang, X., J.~Clune, and K.~O. Stanley\leavevmode\nopagebreak\newline 2017.
\newblock On the relationship between the {OpenAI} evolution strategy and
  stochastic gradient descent.
\newblock {\em arXiv preprint}.
\newblock \arxiv{1712.06564}.

\bibitem[\protect\astroncite{Zhu et~al.}{2022}]{zhu2022torchdrug}
Zhu, Z., C.~Shi, Z.~Zhang, S.~Liu, M.~Xu, X.~Yuan, Y.~Zhang, J.~Chen, H.~Cai,
  J.~Lu, C.~Ma, R.~Liu, L.-P. Xhonneux, M.~Qu, and
  J.~Tang\leavevmode\nopagebreak\newline 2022.
\newblock Torchdrug: A powerful and flexible machine learning platform for drug
  discovery.
\newblock {\em arXiv preprint}.
\newblock \arxiv{2202.08320}.

\end{thebibliography}

\end{document}